\documentclass{article}

\usepackage[nonatbib,preprint]{neurips_2020}

\usepackage[utf8]{inputenc} 
\usepackage[T1]{fontenc}    
\usepackage{hyperref}       
\usepackage{url}            
\usepackage{booktabs}       
\usepackage{amsfonts}       
\usepackage{nicefrac}       
\usepackage{microtype}      

\usepackage{amsmath}
\usepackage{amsthm}
\usepackage{bm}
\usepackage[english]{babel}
\usepackage[backend=biber,style=alphabetic,sorting=ynt]{biblatex}
\usepackage[group-separator={,},group-minimum-digits={3}]{siunitx}
\usepackage{csquotes}
\usepackage{graphicx} 
\usepackage{subcaption}
\usepackage{enumitem}
\usepackage[capitalise]{cleveref}

\DeclareMathOperator*{\argmax}{arg\,max}

\DeclareMathOperator{\diag}{diag}

\DeclareMathOperator{\tr}{tr}
\DeclareMathOperator{\cov}{Cov}
\DeclareMathOperator{\var}{Var}
\DeclareMathOperator{\cv}{CV}
\DeclareMathOperator{\e}{E}

\newcommand*{\estimates}{\mathrel{\hat{=}}}

\renewcommand*{\det}[1]{|#1|}
\newcommand*{\bigdet}[1]{\left|#1\right|}
\newcommand*{\pdet}[1]{|#1|_+}
\newcommand*{\bigpdet}[1]{\left|#1\right|_+}
\newcommand*{\inv}{{{\hbox{-}}1\!}}
\newcommand*{\invsq}{{{\hbox{-}}2}}
\newcommand*{\invsqrt}{{{\hbox{-}}\nicefrac{1}{2}}}
\newcommand*{\tran}{{\mathsf{T}}}

\newcommand*{\itransqrt}{{{\hbox{-}}\nicefrac{\mathsf{T}}{2}}}

\newcommand*{\pder}[2]{\tfrac{\partial#1}{\partial#2}}

\newcommand*{\nicepder}[2]{\nicefrac{\partial#1}{\partial#2}}

\newcommand*{\icol}[1]{
	\left(\begin{smallmatrix}#1\end{smallmatrix}\right)%
}

\newcommand*{\Mu}{\mathrm{M}}
\newcommand*{\Tau}{\mathrm{T}}

\newlist{steps}{enumerate}{1}
\setlist[steps, 1]{label = Step \arabic*:}

\newcounter{eqinline}

\crefname{eqinline}{eq.}{eqs. }
\Crefname{eqinline}{Eq.}{Eqs. }

\title{There and Back Again:\\Unraveling the Variational Auto-Encoder}

\author{%
  Graham Fyffe\thanks{The first version of this work was uploaded while at Google; the text has since been updated.}\\
  \texttt{gfyffe@gmail.com}
}

\begin{document}

\maketitle

\vspace{-1em}

\begin{abstract}
We prove that the evidence lower bound (ELBO) employed by variational auto-encoders (VAEs) admits non-trivial solutions having constant posterior variances under certain mild conditions, removing the need to learn variances in the encoder. The proof follows from an unexpected journey through an array of topics: the closed form optimal decoder for Gaussian VAEs, a proof that the decoder is always smooth, a proof that the ELBO at its stationary points is equal to the exact log evidence, and the posterior variance is merely part of a stochastic estimator of the decoder Hessian.
The penalty incurred from using a constant posterior variance is small under mild conditions, and otherwise discourages large variations in the decoder Hessian.
From here we derive a simplified formulation of the ELBO as an expectation over a batch, which we call the \emph{Batch Information Lower Bound} (BILBO). Despite the use of Gaussians, our analysis is broadly applicable --- it extends to any likelihood function that induces a Riemannian metric. Regarding learned likelihoods, we show that the ELBO is optimal in the limit as the likelihood variances approach zero, where it is equivalent to the change of variables formulation employed in normalizing flow networks. Standard optimization procedures are unstable in this limit, so we propose a bounded Gaussian likelihood that is invariant to the scale of the data using a measure of the aggregate information in a batch, which we call \emph{Bounded Aggregate Information Sampling} (BAGGINS). Combining the two formulations, we construct VAE networks with only half the outputs of ordinary VAEs (no learned variances), yielding improved ELBO scores and scale invariance in experiments. As we perform our analyses irrespective of any particular network architecture, our reformulations may apply to any VAE implementation.
\end{abstract}

\section{Introduction}
\vspace{-0.3ex}

\begin{figure}[h]
\centering
\begin{subfigure}[h]{2.25in}
\includegraphics*[page=1,viewport=0 0 3in 1.5in,clip,scale=0.75]{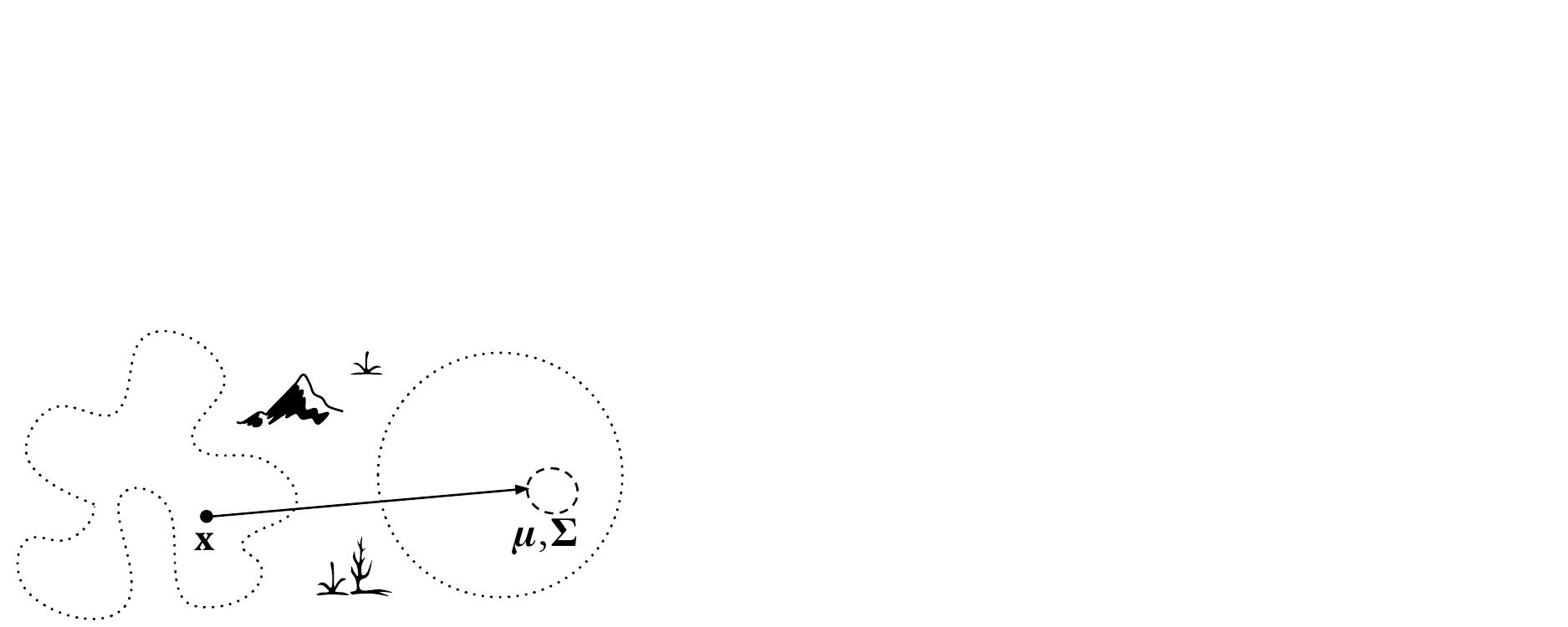}
\caption{There.}
\end{subfigure}
\hfil
\begin{subfigure}[h]{2.25in}
\includegraphics*[page=2,viewport=0 0 3in 1.5in,clip,scale=0.75]{figures/Figures.pdf}
\caption{Back again.}
\end{subfigure}
\caption{
Schematic of a Variational Auto-Encoder. (a) The encoder maps a data vector $\mathbf{x}$ to a posterior distribution (dashed ellipse) parameterized by mean $\bm{\mu}$ and variance $\bm{\Sigma}$ in latent space. (b) The decoder maps a latent vector $\mathbf{z}$ to a likelihood function (dashed ellipse) parameterized by mean $\bm{\nu}$ and variance $\bm{\Tau}$ in data space. The dotted shape and circle represent the data distribution and prior.
}\label{fig-there-and-back-again}
\end{figure}

The Variational Auto-Encoder (VAE) is a powerful model for data analysis, which may be viewed as a more principled formulation of an information bottleneck \cite{2013arXiv1312.6114K}. The underlying idea is that a vector from a population of data with a complex distribution may be the product of some unknown process operating on a latent vector from a much simpler distribution, and if we can model this process we open up a variety of statistical inference applications including dimensionality reduction and novel example generation. These distributions may be estimated by learning an \emph{encoder} which maps data vectors to a latent space, and a \emph{decoder} which maps them back. A VAE treats encoding and decoding as probabilistic operations: a given data vector is encoded to a \emph{posterior distribution} over the latent space (influenced by a \emph{prior}), and a given latent vector is decoded to a \emph{likelihood function} over the data space (\cref{fig-there-and-back-again}). The encoder and decoder are learned together by maximizing the \emph{evidence lower bound} (ELBO) of the marginal likelihood of the data.

In the original VAE formulation, the outputs of the learned encoder include the posterior means and variances, and the outputs of the decoder include the likelihood means and variances. It has been noted, however, that the most common practise in the wild is to employ constant predetermined likelihood variances, and there is good reason --- it has been shown that without regularization, likelihood functions with learned posteriors are unbounded \cite{pierrealex2018leveraging}. Yet when we look at the posterior, learned variances remain the norm. Recently though, works have appeared which also argue for the use of constant posterior variances \cite{Ghoshetal19}. However, to our knowledge no theoretical justification is provided. In the present work, we complete this story. We prove that the evidence lower bound employed by variational auto-encoders admits non-trivial solutions having constant posterior variances under mild conditions, eliminating the need to learn variances in the encoder.

Our proof visits a series of supporting topics, organized into \cref{sec-unexpected}. The proof applies to Gaussian VAEs and extends to any likelihood function that induces a Riemannian metric.
In \cref{sec-solutions} we produce the functional derivative of the ELBO with respect to the decoder, which leads to a closed form solution for the optimal decoder. We then prove that the optimal decoder is always smooth to at least second order, windowed by the posterior --- a property that is often assumed in the VAE literature without proof. We leverage this to prove that the posterior means image the data residuals via the decoder Jacobian, and to produce the closed form posterior variance in terms of the prior and decoder Jacobian.
In \cref{sec-noisy} we construct an exact log evidence formulation from similar assumptions as the ELBO: a latent random process, a mapping to data space, and a data noise model induced by a likelihood function. We prove that this formulation has the same stationary points as the ELBO, but notably it lacks any posterior variances, indicating that those may be ignored when analyzing the other parameters.
In \cref{sec-constant-sigma} we draw from all of the aforementioned results to show that forcing all posterior variances to be equal imposes only a penalty term discouraging large variations in the decoder Hessian, which we argue is not undesirable. Further, we show that under mild conditions (i.e.\ low intrinsic rank and modest relative variation of decoder Hessian) the penalty is small in comparison to the overall ELBO, which completes our main claim.
In \cref{sec-bilbo}, we exploit constant posterior variances to derive a simplified formulation of the ELBO as an expectation over a batch, which we call the \emph{Batch Information Lower Bound} (BILBO).

No analysis of the ELBO is complete without addressing the unboundedness of learned likelihood functions. In \cref{sec-closed-likelihood} we show that the ELBO is optimal in the limit as the likelihood variances approach zero, where it is equivalent to the change of variables formulation employed in normalizing flow networks.
We confirm that standard optimization procedures are unstable in this limit, necessitating further regularization of the likelihoods. We propose a bounded Gaussian likelihood that is invariant to the scale of the data using a measure of the aggregate information in a batch, which we call \emph{Bounded Aggregate Information Sampling} (BAGGINS).
Combining the two proposed formulations, we demonstrate improved evidence lower bounds and scale invariance via experiment.

\paragraph{Contributions}
Our primary contribution is a proof that the ELBO admits non-trivial solutions having constant posterior variances under certain mild conditions, eliminating the need to learn variances in the encoder. Our supporting contributions include:

\begin{itemize}
  \item Derivation of the closed form optimal decoder for Gaussian VAEs.
  \item Proof of the smoothness of the optimal decoder.
  \item An exact log evidence formulation connecting VAEs and normalizing flow networks.
  \item A simplification of ELBO exploiting constant posterior variances, called BILBO.
  \item A scale-invariant bounding scheme for Gaussian likelihoods, called BAGGINS.
\end{itemize}

\section{Related Work}

The VAE literature is thriving, with works exploring a multitude of applications, network architectures, and enhancements. For example, some works combine VAEs with normalizing flows to enable richer posteriors \cite{pmlr-v37-rezende15,NIPS2016_6581} or decoders \cite{morrow2020variational}. Our work is most related to those that go back to the original formulation of the ELBO objective \cite{2013arXiv1312.6114K} to look for new insights. Aspects under analysis include disentanglement \cite{Higgins2017betaVAELB,burgess2018understanding}, convergence and rate-distortion characteristics \cite{Alemi2018FixingAB}, robustness to outliers and interpretation of posterior variance eigenvalues \cite{dai2017hidden}, and the impact of Gaussian posteriors and likelihoods on the effectiveness of the VAE \cite{dai2019diagnosing}.

\section{Preliminaries}\label{sec-preliminaries}

\paragraph{Variational Auto-Encoders \cite{2013arXiv1312.6114K}}
We assume an observed value $\mathbf{x} \sim p(\mathbf{x})$ results from an unobserved or latent variable $\mathbf{z} \sim p(\mathbf{z})$, typically of lower dimension, via a conditional distribution (or likelihood) $p(\mathbf{x}|\mathbf{z})$. We are interested in the posterior $p(\mathbf{z}|\mathbf{x})$, which is in general intractable. We estimate a tractable prior distribution $p_{\bm{\theta}}(\mathbf{z})$ and likelihood $p_{\bm{\theta}}(\mathbf{x}|\mathbf{z})$, parameterized by $\bm{\theta}$. Further, we estimate an approximate posterior $q_{\bm{\phi}}(\mathbf{z}|\mathbf{x})$, parameterized by $\bm{\phi}$, rather than the intractable true posterior.
The likelihood $p_{\bm{\theta}}(\mathbf{x}|\mathbf{z})$ is often referred to interchangeably as a probabilistic decoder, and the approximate posterior $q_{\bm{\phi}}(\mathbf{z}|\mathbf{x})$ as a probabilistic encoder, as these are the roles they play in mapping values from latent space to data space and back. Correspondingly, $\bm{\theta}$ may be referred to as the generative parameters, and $\bm{\phi}$ as the recognition parameters.
We estimate these parameters by maximizing the so-called variational lower bound, or evidence lower bound (ELBO):
\begin{equation}\label{eq-vae}
    \bm{\theta}^*, \bm{\phi}^* = \argmax_{\bm{\theta}, \bm{\phi}}
    \e_{p(\mathbf{x})} [\text{ELBO}]; \;
        \text{ELBO} \equiv -\mathrm{KL}(q_{\bm{\phi}}(\mathbf{z}|\mathbf{x}) | p_{\bm{\theta}}(\mathbf{z}))
        + \e_{q_{\bm{\phi}}(\mathbf{z}|\mathbf{x})} \left[
            \log p_{\bm{\theta}}(\mathbf{x}|\mathbf{z})
        \right],
\end{equation}
where $\mathrm{KL}(q | p)$ is the Kullback-Leibler divergence (or relative entropy) of $q$ with respect to $p$, equal to
$\e_{q(\mathbf{z})} \left[
        \log q(\mathbf{z}) - \log p(\mathbf{z}))
    \right]$.
Intuitively, the KL term encourages the posterior to resemble the prior, while the log likelihood term encourages maximum likelihood.

\paragraph{Map Notation}
The approximate posterior may be viewed as a mapping $\bm{\mu_\phi} : \mathbb{R}^m \rightarrow \mathbb{R}^n$ that maps a data vector to the mean of its distribution in latent space, and an additional mapping for the other parameters of its distribution in latent space (e.g.\ variance). Similarly, the likelihood may be viewed as a mapping $\bm{\nu_\theta} : \mathbb{R}^n \rightarrow \mathbb{R}^m$ that maps a latent vector to the mean of its distribution in data space, and an additional mapping for the other parameters of its distribution in data space (e.g.\ variance). In the following, we make references to these mappings, often dropping their arguments for clarity.

\paragraph{Gaussian Variational Auto-Encoders}
We consider a variational auto-encoder with Gaussian approximate posterior 
$q_{\bm{\phi}}(\mathbf{z}|\mathbf{x})\!=\!\mathcal{N}(\mathbf{z}; \bm{\mu_\phi}(\mathbf{x}), \bm{\Sigma_\phi}(\mathbf{x}))$ and Gaussian prior $p_{\bm{\theta}}(\mathbf{z}) = \mathcal{N}(\mathbf{z}; \mathbf{0}, \mathbf{S}^2)$, with diagonal $\bm{\Sigma}$ and $\mathbf{S}$.
Dropping the $\!_{\bm{\phi}\!}(\mathbf{x})$ parameters for clarity, \cref{eq-vae} becomes:
\begin{equation}\label{eq-vae-s}
\text{ELBO} =
\tfrac{1}{2} \! \left(
\log \det{e \mathbf{S}^\invsq \bm{\Sigma}}
- \tr \mathbf{S}^\invsq \bm{\Sigma}
- |\mathbf{S}^\inv \bm{\mu}|^2 \right)
+ \e_{\mathcal{N}(\mathbf{z}; \bm{\mu}, \bm{\Sigma})} [
\log p_{\bm{\theta}}(\mathbf{x}|\mathbf{z}) ].
\end{equation}

\section{An Unexpected Journey}\label{sec-unexpected}

In this section, we prove that the ELBO admits non-trivial solutions having constant posterior variances under mild conditions. As outlined in the Introduction, we characterize the first order optimality conditions of the ELBO for Gaussian VAEs, relate the stationary points to a log evidence formulation having no posterior variances, and finally bring it all together to prove the main claim. We follow up with a discussion of implications of the proof, including a simplification of the ELBO.
 
\subsection{The First Order Optimality Conditions of the ELBO}\label{sec-solutions}

We characterize the first order optimality conditions of the ELBO assuming Gaussian prior and posterior. We were unable to find the partial derivatives of the ELBO in the literature, so we present them here. Each partial derivative provides a clue about what the VAE is doing, in the form of an optimality condition relating some of its parameters. Note that we place no restrictions on the encoder or decoder, which may be general nonlinear functions. We consider unit normal likelihoods $p_{\bm{\theta}}(\mathbf{x}|\mathbf{z}) = \mathcal{N}(\bm{\nu}(\mathbf{z}); \mathbf{x},\mathbf{I})$, which induce a Euclidean metric on the data space. Our analysis, however, extends to any likelihood function that induces a Riemannian metric on the data space.

\paragraph{Proof of Extension}
Any Riemannian space may be isometrically embedded into a Euclidean space (per the Nash embedding theorem), where geodesic distances may be locally approximated by Euclidean distances within a region $p_{\bm{\theta}}(\mathbf{x}) > \epsilon$ through the introduction of additional dimensions to mitigate self intersection \cite{10.5555/2567709.2567739}. As all of the variation in the data population is contained within the embedded manifold, none of the latent dimensions learned by an auto-encoder would be tasked with representing the newly added dimensions, and therefore the latent vectors may remain unchanged.
Henceforth, $\mathbf{x}$ represents a data point conceptually embedded into a Euclidean space, but importantly, our results do not require computing such an embedding, and merely rely on its theoretical existence.

\paragraph{The Prior}
The partial derivative of $\e_{p(\mathbf{x})} [\text{ELBO}]$ with respect to $\mathbf{S}^\invsq$ (rather than $\mathbf{S}$) is:
\begin{align}
\pder{}{\mathbf{S}^\invsq} \e_{p(\mathbf{x})} [\text{ELBO}]
= \tfrac{1}{2} \e_{p(\mathbf{x})} [\mathbf{S}^2 - \bm{\Sigma} - \bm{\mu}\bm{\mu}^\tran].
\label{eq-partial-s}
\end{align}
Traditionally the prior is a given, but here we solve for it in terms of the posteriors.
Solving $\nicepder{}{\mathbf{S}^\invsq} \e_{p(\mathbf{x})} [\text{ELBO}] = \mathbf{0}^\tran$ with \cref{eq-partial-s} admits a closed form solution for $\mathbf{S}^2$ in terms of the posteriors:
\begin{equation}
	\mathbf{S}^2 = \e_{p(\mathbf{x})} [\bm{\Sigma} + \bm{\mu}\bm{\mu}^\tran].
\label{eq-closed-s}
\end{equation}

\paragraph{The Decoder}
The functional derivative of $\e_{p(\mathbf{x})} [\text{ELBO}]$ with respect to the decoder $\bm{\nu}(\mathbf{z})$ is:
\begin{align}
\pder{}{\bm{\nu}(\mathbf{z})}
\e_{p(\mathbf{x})} [\text{ELBO}]
=
-\tfrac{1}{N} \sum_i \int_{\mathbf{z}}\!
\mathcal{N}(\mathbf{z}; \bm{\mu}_i, \bm{\Sigma}_i)
(\bm{\nu}(\mathbf{z}) - \mathbf{x}_i)^\tran d\mathbf{z};
\label{eq-partial-nu}
\end{align}
where $N$ is the number of data examples in the population.
Solving $\nicepder{}{\bm{\nu}(\mathbf{z})} \e_{p(\mathbf{x})} [\text{ELBO}] = \mathbf{0}^\tran$ with \cref{eq-partial-nu} admits a closed form solution for $\bm{\nu}(\mathbf{z})$ in terms of the posteriors:
\begin{equation}
\bm{\nu}(\mathbf{z}) =
\frac{\sum_i \mathcal{N}(\mathbf{z}; \bm{\mu}_i, \bm{\Sigma}_i) \mathbf{x}_i }
{\sum_i \mathcal{N}(\mathbf{z}; \bm{\mu}_i, \bm{\Sigma}_i)}.
\label{eq-closed-warp}
\end{equation}
We also derive its Jacobian, referring to \cite{IMM2012-03274}\footnote{\cite{IMM2012-03274} Eq.~(347) contains an error; the expression is transposed.}:
\begin{equation}
\pder{}{\mathbf{z}}\bm{\nu}(\mathbf{z})
=
\frac{\sum_i \mathcal{N}(\mathbf{z}; \bm{\mu}_i, \bm{\Sigma}_i)
	(\bm{\nu}(\mathbf{z}) - \mathbf{x}_i) (\mathbf{z} - \bm{\mu}_i)^\tran \bm{\Sigma}_i^\inv}
{\sum_i \mathcal{N}(\mathbf{z}; \bm{\mu}_i, \bm{\Sigma}_i)}.
\label{eq-closed-warp-jacobian}
\end{equation}
In the following we let $\mathbf{J}$ denote the Jacobian of the decoder evaluated at the encoded vector:
\begin{equation}
\mathbf{J} \equiv \pder{}{\mathbf{z}}\bm{\nu}(\mathbf{z}) \big|_{\mathbf{z}=\bm{\mu}}.
\label{eq-j-definition}
\end{equation}

\paragraph{Proof of the Smoothness of the Optimal Decoder}
On the mild condition that $\bm{\Sigma}_i \simeq \bm{\Sigma}_j$ whenever $\mathcal{N}(\bm{\mu}_j; \bm{\mu}_i, \bm{\Sigma}_i)$ and $\mathcal{N}(\bm{\mu}_i; \bm{\mu}_j, \bm{\Sigma}_j)$ are significant, we recognize $\mathbf{J}$ as moving least squares linear regression windowed by the posterior, with the outer product term estimating $\cov(\mathbf{x}, \bm{\mu})$ and $\bm{\Sigma}$ standing in for $\var(\bm{\mu})$.
This implies that the decoder is smooth over the posterior $\mathcal{N}(\mathbf{z}; \bm{\mu}, \bm{\Sigma})$.

\paragraph{The Posteriors}
We have obtained solutions for the prior and decoder in closed form in terms of the posteriors. Therefore the prior and decoder may be substituted by their solutions to aid in analyzing the posteriors.
The partial derivatives of $\e_{p(\mathbf{x})} [\text{ELBO}]$ with respect to $\bm{\mu}$ and $\bm{\Sigma}$ are as follows (using the partial derivatives of $\e_{\mathcal{N}(\mathbf{z}; \bm{\mu}, \bm{\Sigma})} [\log p_{\bm{\theta}}(\mathbf{x}|\mathbf{z})]$ detailed in \cref{appendix-dmu,appendix-dsigma}):
\begin{align}
\pder{}{\bm{\mu}} \e_{p(\mathbf{x})} [ \text{ELBO} ]
\simeq
-\tfrac{1}{N} \big( \bm{\mu}^\tran \mathbf{S}^\invsq
+ (\bm{\nu}(\bm{\mu}) - \mathbf{x})^\tran \mathbf{J} \big);
\label{eq-partial-mu}
\\
\pder{}{\bm{\Sigma}} \e_{p(\mathbf{x})} [ \text{ELBO} ]
\simeq
\tfrac{1}{2N} \big( \bm{\Sigma}^\inv - \mathbf{S}^\invsq - \mathbf{J}^\tran\! \mathbf{J} \big).
\label{eq-partial-sigma}
\end{align}
The errors in the approximations denoted by $\simeq$ in \cref{eq-partial-mu,eq-partial-sigma} are due to treating the mapping $\bm{\nu}$ as locally linear when computing expectations over the posterior. These errors are always small: recall that \cref{eq-closed-warp-jacobian} implies not only that the optimal mapping $\bm{\nu}(\mathbf{z})$ is smooth over the posterior, but its Jacobian is equivalent to a moving least squares linear regression model windowed by the posterior. 
Solving $\nicepder{}{\bm{\mu}} \e_{p(\mathbf{x})} [ \text{ELBO} ] = \mathbf{0}^\tran$ with \cref{eq-partial-mu} implies that the posterior means image the residuals:
\begin{align}
\bm{\mu} \simeq \mathbf{S}^2 \mathbf{J}^\tran\! (\mathbf{x} - \bm{\nu}(\bm{\mu})).
\label{eq-closed-mu}
\end{align}
Solving $\nicepder{}{\bm{\Sigma}} \e_{p(\mathbf{x})} [ \text{ELBO} ] = \mathbf{0}^\tran$ with \cref{eq-partial-sigma} implies that the posterior precision is equal to the prior precision plus the information (the Hessian of the negative log likelihood):
\begin{align}
\bm{\Sigma}^\inv \simeq \mathbf{S}^\invsq + \mathbf{J}^\tran\! \mathbf{J}.
\label{eq-closed-sigma}
\end{align}
\cref{eq-closed-mu,eq-closed-sigma} are not closed form solutions due to their recursive nature, but are useful equations expressing the first order optimality conditions of the ELBO, and will be employed in the following.

\subsection{The Exact Log Evidence}\label{sec-noisy}

The ELBO is established as a lower bound on log evidence through the use of Jensen's inequality and KL divergence \cite{2013arXiv1312.6114K}. We prove that it is in fact equal to exact log evidence at its stationary points to at least a second order approximation, and thus the optimal ELBO is a tight bound.
As we know of no derivation of the exact log evidence in the literature, we produce it here.
Let the data vector $\mathbf{x}$ be modeled in terms of a random vector $\mathbf{y} = \icol{\bm{\zeta} \\ \bm{\gamma}}$ consisting of an underlying latent process $\bm{\zeta}$ and a data noise process $\bm{\gamma}$, as $\mathbf{x} = g(f(\bm{\zeta}), \bm{\gamma})$ with differentiable functions $f : \mathbb{R}^n \rightarrow \mathbb{R}^m$ and $g : \mathbb{R}^m \times \mathbb{R}^m \rightarrow \mathbb{R}^m$.
We augment $\mathbf{x}$ with $\epsilon \bm{\zeta}$ as $\mathbf{\hat{x}} = \icol{\mathbf{x} \\ \epsilon \bm{\zeta}}$ to produce an arbitrary diffeomorphism allowing the log probability density to be calculated via a change of variables:
\begin{equation}
\log p_{\bm{\theta}}(\mathbf{\hat{x}})
= \log p(\mathbf{y}) - \log \bigpdet{\pder{\mathbf{\hat{x}}}{\mathbf{y}}},
\end{equation}
where $\pdet{\circ}$ represents pseudo-determinant \cite{Holbrook:2018:DPD} which allows us to consider the degenerate distribution in the limit $\epsilon \rightarrow 0$ in order to collapse the extra dimensions:
\begin{equation}
\lim_{\epsilon \rightarrow 0}
\log p_{\bm{\theta}}(\mathbf{\hat{x}})
= \log p(\mathbf{y})
- \tfrac{1}{2} \log \bigdet{
    \pder{\mathbf{x}}{\bm{\zeta}} \pder{\mathbf{x}}{\bm{\zeta}}^{\!\tran} +	\pder{\mathbf{x}}{\bm{\gamma}} \pder{\mathbf{x}}{\bm{\gamma}}^{\!\tran}
}.
\end{equation}
This may be expanded and rearranged with the log evidence on the left hand side:
\begin{equation}
\log p_{\bm{\theta}}(\mathbf{x})
= - \log \lim_{\epsilon \rightarrow 0} p_{\bm{\theta}}(\epsilon \bm{\zeta} | \mathbf{x})
 + \log p(\bm{\zeta})
 + \log p(\bm{\gamma})
 - \tfrac{1}{2} \log \bigdet{
    \pder{\mathbf{x}}{\bm{\zeta}} \pder{\mathbf{x}}{\bm{\zeta}}^{\!\tran} +	\pder{\mathbf{x}}{\bm{\gamma}} \pder{\mathbf{x}}{\bm{\gamma}}^{\!\tran}
}.
\label{eq-exact-noisy-general}
\end{equation}
\cref{eq-exact-noisy-general} is the exact log evidence for the general model, which is well defined despite the awkward degenerate limit in the right hand side. Consider the case of Gaussian distributions employed in earlier sections.
Let $\bm{\zeta} = \mathbf{S}^\inv\mathbf{z}$ and $\bm{\gamma} = \mathbf{x} - \bm{\nu}(\mathbf{z})$, noting that the prior and unit normal likelihood induce standard normal distributions on both. Then model $\mathbf{x}$ as $\mathbf{x} = \bm{\nu}(\mathbf{S}\bm{\zeta}) + \bm{\gamma}$, yielding:
\begin{align}
\log p_{\bm{\theta}}(\mathbf{x})
=
- \tfrac{1}{2} \big(
m \log 2 \pi
+ |\mathbf{S}^\inv\mathbf{z}|^2
+ |\mathbf{x} - \bm{\nu}(\mathbf{z})|^2
+ \log \bigdet{\pder{\bm{\nu}}{\bm{z}}^{\!\tran}\!\pder{\bm{\nu}}{\bm{z}} \mathbf{S}^2 + \mathbf{I}}
\big),
\label{eq-exact-noisy}
\end{align}
where the degenerate limit term cancels the $-\nicefrac{n}{2}\log 2\pi$ term in $\log p(\bm{\zeta})$.
\cref{eq-exact-noisy} is the exact log evidence for standard normal data noise. If the decoder Jacobian $\nicepder{\bm{\nu}}{\bm{z}}$ is available, one may maximize the exact log evidence rather than the ELBO, in much the same way that normalizing flow networks are trained by maximizing the exact log likelihood using network layers with tractable Jacobians \cite{dinh2017density}.
Supposing we were to do this, how would the result compare to using the ELBO?
To make the comparison easier, we rewrite the ELBO using $\e_{\mathcal{N}(\mathbf{z}; \bm{\mu}, \bm{\Sigma})}[|\mathbf{x} - \bm{\nu}(\mathbf{z})|^2] \simeq |\mathbf{x} - \bm{\nu}(\bm{\mu})|^2 + \tr \mathbf{J}^\tran\!\mathbf{J}\bm{\Sigma}$, which holds to at least second order accuracy due to the smoothness of the decoder per \cref{eq-closed-warp-jacobian}:
\begin{align}
\text{ELBO} \simeq
- \tfrac{1}{2} \big(
m \log 2 \pi
+ |\mathbf{S}^\inv\bm{\mu}|^2
+ |\mathbf{x} - \bm{\nu}(\bm{\mu})|^2
+ \tr ((\mathbf{S}^\invsq + \mathbf{J}^\tran\!\mathbf{J}) \bm{\Sigma})
- \log \det{e \mathbf{S}^\invsq \bm{\Sigma}}
\big).
\label{eq-elbo-like-noisy}
\end{align}
We examine the difference between \cref{eq-exact-noisy} and  \cref{eq-elbo-like-noisy} at $\mathbf{z} = \bm{\mu}$, collecting and canceling terms:
\begin{align}
\log p_{\bm{\theta}}(\mathbf{x}) - \text{ELBO}
\simeq
\tfrac{1}{2} \big(
\tr ((\mathbf{S}^\invsq + \mathbf{J}^\tran\!\mathbf{J}) \bm{\Sigma})
- \log \det{e (\mathbf{S}^\invsq + \mathbf{J}^\tran\!\mathbf{J}) \bm{\Sigma}}
\big).
\label{eq-elbo-diff}
\end{align}
Clearly, \cref{eq-elbo-diff} is zero when $\bm{\Sigma} = (\mathbf{S}^\invsq + \mathbf{J}^\tran\!\mathbf{J})^\inv$. As this agrees with the solution from \cref{eq-closed-sigma}, we conclude that the ELBO is equal to the exact log evidence at its stationary points.

\paragraph{The Role of $\bm{\Sigma}$}
Intriguingly, $\bm{\Sigma}$ may be viewed as part of a stochastic estimator for the log determinant of Jacobian term in \cref{eq-exact-noisy} using only traces, which are easily obtained by probing with random vectors \cite{avron2011}. Consider the following equation, which holds for any Hermitian matrix $\mathbf{A}$:
\begin{align}
\min_{\bm{\Sigma}} \tr \mathbf{A} \bm{\Sigma} - \log \det{e \bm{\Sigma}}
= \log \det{\mathbf{A}}.
\label{eq-trace-log}
\end{align}
This allows the log determinant of $\mathbf{A}$ to be computed using e.g.\ stochastic gradient descent over $\bm{\Sigma}$ by probing $\mathbf{v}^\tran\!\mathbf{A} \mathbf{v}$ with vectors $\mathbf{v} \sim \mathcal{N}(\bm{0}, \bm{\Sigma})$. This is precisely what happens with the ELBO:
\begin{align}
\min_{\bm{\Sigma}} \tr ((\mathbf{S}^\invsq + \mathbf{J}^\tran\!\mathbf{J}) \bm{\Sigma})
- \log \det{e \mathbf{S}^\invsq \bm{\Sigma}}
= \log \det{\mathbf{J}^\tran\!\mathbf{J} \mathbf{S}^2 + \mathbf{I}}.
\label{eq-sigma-role}
\end{align}
Viewed in this light, it becomes clear that $\bm{\Sigma}$ is not fundamental to the solution of the optimal model parameters for \cref{eq-exact-noisy}, which exist regardless of whether we compute the Jacobian directly, or use the stochastic scheme with \cref{eq-sigma-role} i.e.\ the ELBO. We will recall this point in the following sections.

\subsection{The Case for Constant Posterior Variances}\label{sec-constant-sigma}

Previous works argue for constant entropy posteriors by intuition and experiment \cite{Ghoshetal19}. Here, we provide a principled analysis.
Examining \cref{eq-elbo-diff} more closely, we find an intriguing pattern --- the trace term conceals a first order approximation of the log term, and they largely cancel:
\begin{align}
\log p_{\bm{\theta}}(\mathbf{x}) - \text{ELBO}
\simeq
\tfrac{1}{4} \tr^2 ((\mathbf{S}^\invsq + \mathbf{J}^\tran\!\mathbf{J})\bm{\Sigma} - \mathbf{I})
+ O\!\left( \tr^3 ((\mathbf{S}^\invsq + \mathbf{J}^\tran\!\mathbf{J})\bm{\Sigma} - \mathbf{I}) \right).
\label{eq-penalty-taylor}
\end{align}
Thus the value of the ELBO should be somewhat insensitive to the posterior variance, motivating the question: what would happen if we forced all posterior variances to be equal? The optimal such variance is $\bm{\tilde{\Sigma}} = \e_{p(\mathbf{x})} [\mathbf{S}^\invsq + \mathbf{J}^\tran\!\mathbf{J}]^\inv$, which we plug into the expectation of \cref{eq-penalty-taylor} to analyze the penalty due to employing a constant posterior variance (dropping the third-order term):
\begin{align}
\e_{p(\mathbf{x})}\! \left[ \log p_{\bm{\theta}}(\mathbf{x}) - \text{ELBO}_{\bm{\tilde{\Sigma}}} \right]
\simeq
\tfrac{1}{4} \tr \cv^2_{p(\mathbf{x})} [\mathbf{S}^\invsq +  \mathbf{J}^\tran\!\mathbf{J}],
\label{eq-penalty}
\end{align}
where $\cv^2$ represents the squared coefficient of variation, or \emph{relative variance}.
(Since $\mathbf{S}$ is diagonal we may also write the penalty as $\nicefrac{1}{4} \tr \cv^2_{p(\mathbf{x})} [\mathbf{J}^\tran\!\mathbf{J} \mathbf{S}^2 + \mathbf{I}]$ to clarify its relationship to the log determinant in \cref{eq-exact-noisy,eq-sigma-role}.)
The penalty vanishes when the decoder Hessian is homogeneous. Otherwise, it acts as a regularization term to penalize large variations in the decoder Hessian.

\paragraph{Magnitude of the Penalty}
The magnitude of the penalty in \cref{eq-penalty} is determined by both the rank and the coefficient of variation of the decoder Hessian. The rank is the lesser of the latent dimension $n$ and the dimension of the intrinsic data manifold (without noise). This is typically much smaller than the data dimension $m$ in domains where VAEs are applied, such being the underlying assumption motivating the VAE in the first place.
The coefficient of variation on the other hand is not technically bounded. However, for many reasonable distributions it is near unity or smaller. To be clear, we are referring to the distribution of decoder Hessians at the data examples, not the distribution of the data examples themselves.
We may conclude under these mild conditions, that the ELBO in \cref{eq-elbo-like-noisy} is dominated by the residual term with its $m$ dimensions of noise. The penalty is small in comparison.

\begin{figure}[ht]
\begin{subfigure}[h]{0.32\linewidth}
\includegraphics[width=\linewidth]{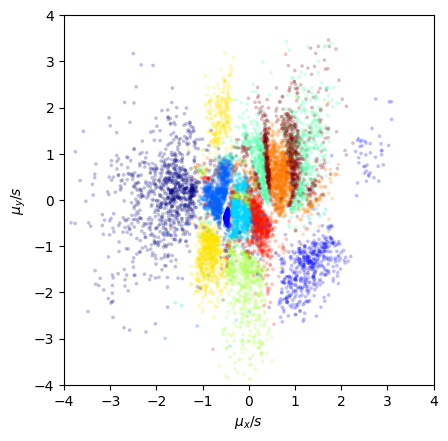}
\caption{Run 1.}
\end{subfigure}
\hfil
\begin{subfigure}[h]{0.32\linewidth}
\includegraphics[width=\linewidth]{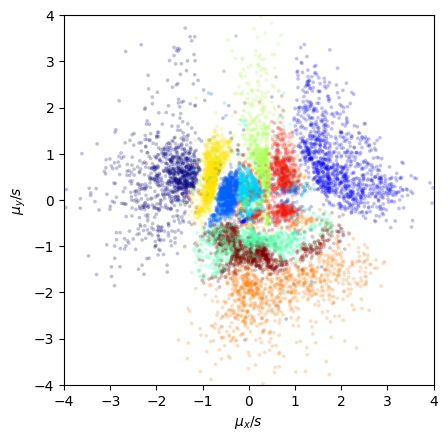}
\caption{Run 2.}
\end{subfigure}
\hfil
\begin{subfigure}[h]{0.32\linewidth}
\includegraphics[width=\linewidth]{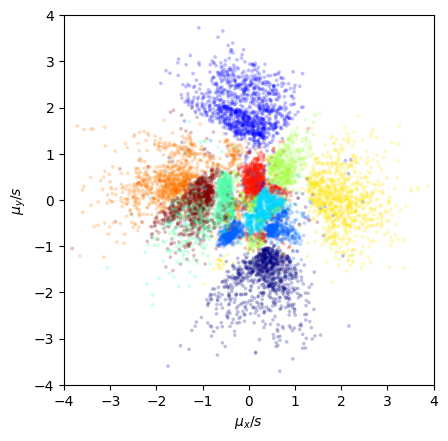}
\caption{Run 3.}
\end{subfigure}
\caption{
Posterior means for a two-dimensional MNIST latent space colored by label, with learned posterior variances and Bernoulli likelihood, for three runs with identical parameters. Different runs produce different clusters because of the stochastic nature of the optimizer, but yield comparable variational lower bounds (here \num{-136.5}, \num{-135.0}, and \num{-134.4}).
}\label{fig-bernoulli-scatter-runs}
\end{figure}

\begin{figure}[ht]
\begin{subfigure}[h]{0.32\linewidth}
\includegraphics[width=\linewidth]{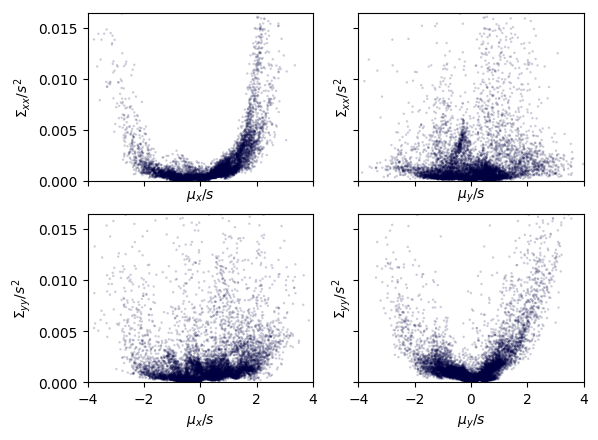}
\caption{Run 1.}
\end{subfigure}
\hfil
\begin{subfigure}[h]{0.32\linewidth}
\includegraphics[width=\linewidth]{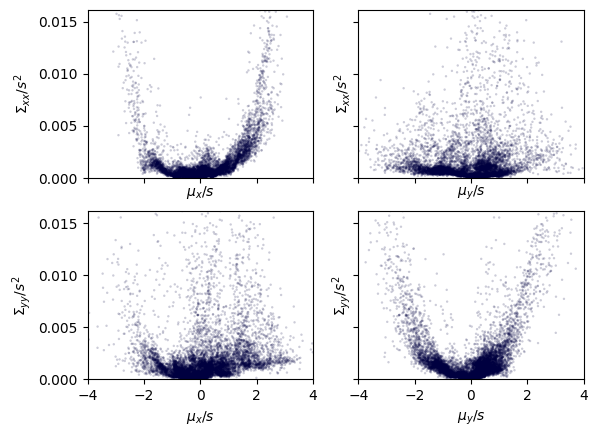}
\caption{Run 2.}
\end{subfigure}
\hfil
\begin{subfigure}[h]{0.32\linewidth}
\includegraphics[width=\linewidth]{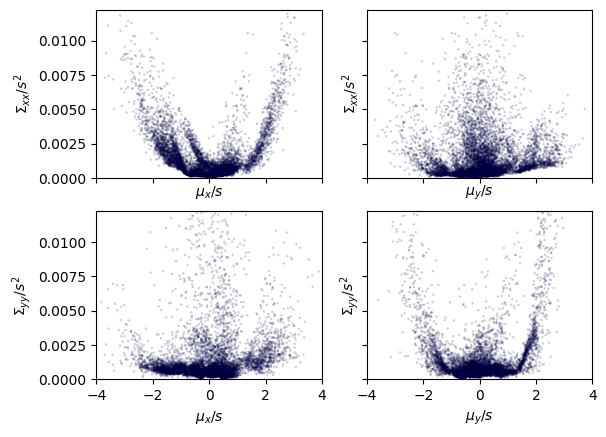}
\caption{Run 3.}
\end{subfigure}
\caption{
Learned posterior variances for the same runs as \cref{fig-bernoulli-scatter-runs}, plotting $x$ and $y$ variances vs.\ mean $x$ and $y$ relative to prior scale $s^2$ (1 in this experiment). Note the U shape when plotted against the same mean dimension, and lack of correlation to the other dimension, consistent with inflating variances near the fringes rather than revealing intrinsic structure in the data.
}\label{fig-variance-u-shape}
\end{figure}

\paragraph{Effects of the Penalty}
Though the penalty is small, we still examine its effects on the decoder. We note that global anisotropic scaling, and locally varying isometries to better align the decoded density to the true density, are all admitted without penalty. This preserves the local PCA behavior of VAEs \cite{vae-pca}.
What is penalized is locally varying scaling.
This discourages arbitrary local contractions or expansions that force a match to the prior at each point in space,
and instead encourages rearranging the intrinsic density of the data population to fit the prior density.
Consider the scatter plots in \cref{fig-bernoulli-scatter-runs,fig-variance-u-shape} illustrating how standard VAEs inflate the posterior variances in the fringes of the latent distribution, despite different training runs producing different arrangements of the data.
We therefore argue that even in cases where the penalty is not small, its effects are not undesirable.

\paragraph{The Floating Prior Trick}\label{sec-floating-prior}
With all posterior variances equal, \cref{eq-closed-s} appears to allow us to determine $\bm{\Sigma}$ simply by plugging in the desired $\mathbf{S}$ and an estimate of $\bm{\Mu} \equiv \e_{p(\mathbf{x})} [\bm{\mu}\bm{\mu}^\tran]$. Unfortunately, during training there is no guarantee that $\bm{\Sigma} = \mathbf{S}^2 - \bm{\Mu}$ is positive definite. We therefore propose to fix $\bm{\Sigma}$ and allow the prior to \emph{float}, via $\mathbf{S}^2 = \bm{\Sigma} + \bm{\Mu}$. We may select any $\bm{\Sigma}$, such as $\bm{\Sigma} = \mathbf{I}$, but we must estimate $\bm{\Mu}$. To obtain a robust estimate in a deep network training setting, we average over a batch $B$ with $\bm{\Mu}_B \estimates \e_B [\diag(\bm{\mu})^2]$ and $\mathbf{S}^2 = \bm{\Sigma} + \bm{\Mu}_B$. Assuming a diagonal $\bm{\Mu}$ does not impact the ELBO, which is invariant to linear latent transformations. Further, the intrinsic latent dimension \cite{dai2017hidden} is captured in $\bm{\Mu}$ anisotropy rather than $\bm{\Sigma}$ anisotropy. We then evaluate \cref{eq-vae-s} as usual.
\cref{fig-elbo-burndown} illustrates that constant posterior variance slightly improves the lower bound on MNIST experiments, and \cref{fig-generative-learned-vs-bilbo,fig-bernoulli-scatter} illustrate that generative samples and posterior distributions are qualitatively similar to those of learned posterior variances. See \cref{sec-experiments} for more details.

\subsection{Batch Information Lower Bound}\label{sec-bilbo}

\begin{figure}[h!]
	\begin{subfigure}[h]{0.3\linewidth}
		\includegraphics[width=\linewidth]{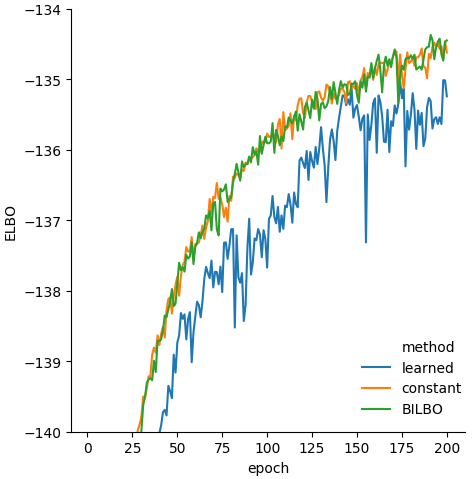}
		\caption{2 latent dimensions.}\label{fig-burndown}
	\end{subfigure}
	\hfil
	\begin{subfigure}[h]{0.3\linewidth}
		\includegraphics[width=\linewidth]{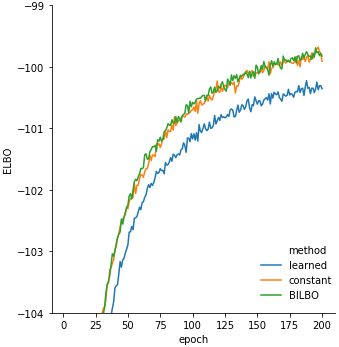}
		\caption{100 latent dimensions.}\label{fig-burndown100}
	\end{subfigure}
	\hfil	
	\begin{subfigure}[h]{0.35\linewidth}
		\vspace{1ex}
		\begin{subfigure}[h]{0.22\linewidth}
			\tiny \quad
		\end{subfigure}
		\begin{subfigure}[h]{0.75\linewidth}
			\centering
			\tiny 2 latent dimensions
		\end{subfigure}\\
		\begin{subfigure}[h]{0.22\linewidth}
			\raggedleft
			\tiny Learned $\bm{\Sigma}$
		\end{subfigure}
		\hfil
		\begin{subfigure}[h]{0.75\linewidth}
			\includegraphics[width=\linewidth]{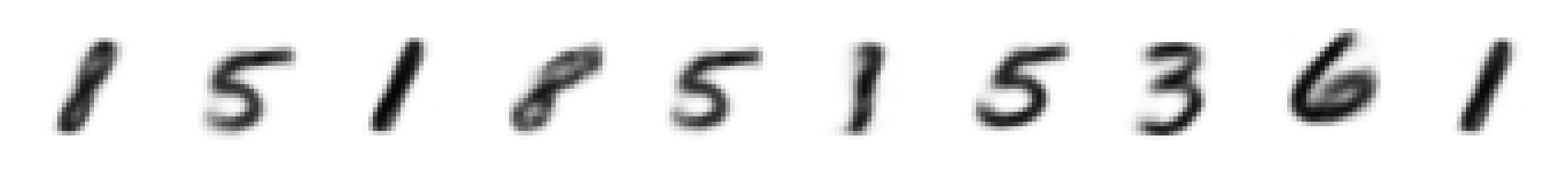}
		\end{subfigure}\\
		\begin{subfigure}[h]{0.22\linewidth}
			\raggedleft
			\tiny Constant $\bm{\Sigma}$
		\end{subfigure}
		\hfil
		\begin{subfigure}[h]{0.75\linewidth}
			\includegraphics[width=\linewidth]{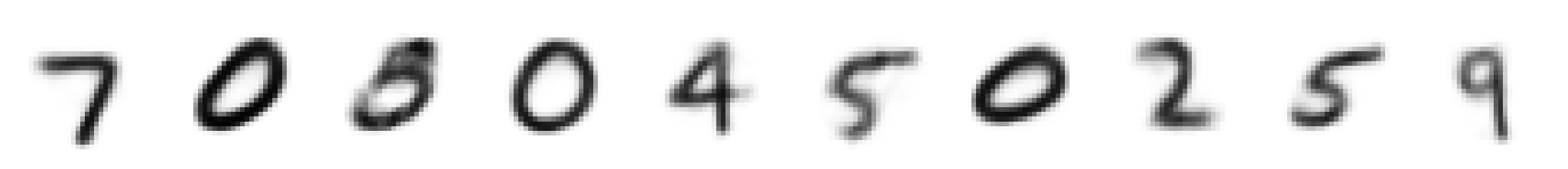}
		\end{subfigure}\\
		\begin{subfigure}[h]{0.22\linewidth}
			\raggedleft
			\tiny BILBO
		\end{subfigure}
		\hfil
		\begin{subfigure}[h]{0.75\linewidth}
			\includegraphics[width=\linewidth]{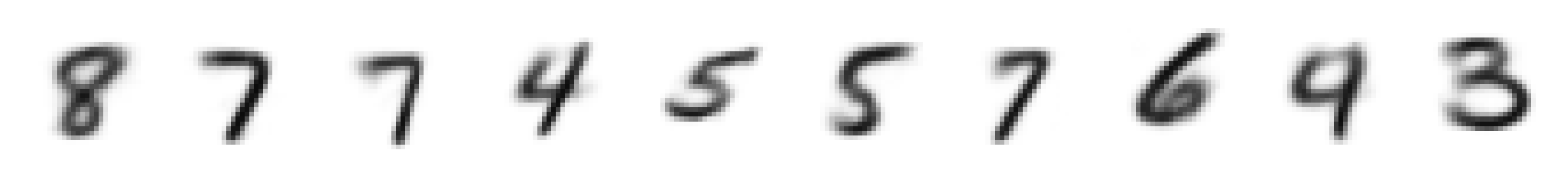}
		\end{subfigure}
		\vspace{1em}\\
		\begin{subfigure}[h]{0.22\linewidth}
			\tiny \quad
		\end{subfigure}
		\begin{subfigure}[h]{0.75\linewidth}
			\centering
			\tiny 100 latent dimensions
		\end{subfigure}\\
		\begin{subfigure}[h]{0.22\linewidth}
			\raggedleft
			\tiny Learned $\bm{\Sigma}$
		\end{subfigure}
		\hfil
		\begin{subfigure}[h]{0.75\linewidth}
			\includegraphics[width=\linewidth]{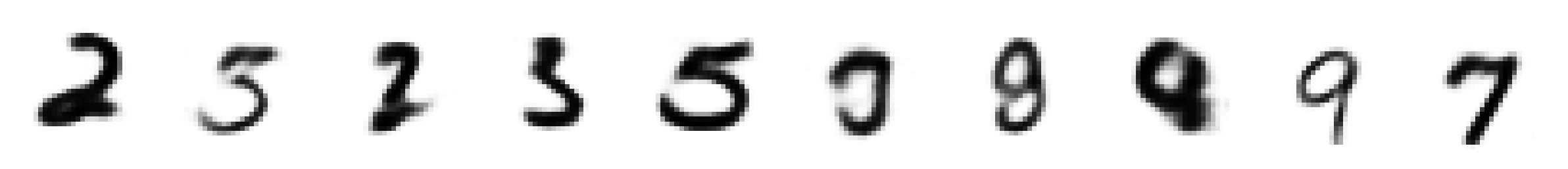}
		\end{subfigure}\\
		\begin{subfigure}[h]{0.22\linewidth}
			\raggedleft
			\tiny Constant $\bm{\Sigma}$
		\end{subfigure}
		\hfil
		\begin{subfigure}[h]{0.75\linewidth}
			\includegraphics[width=\linewidth]{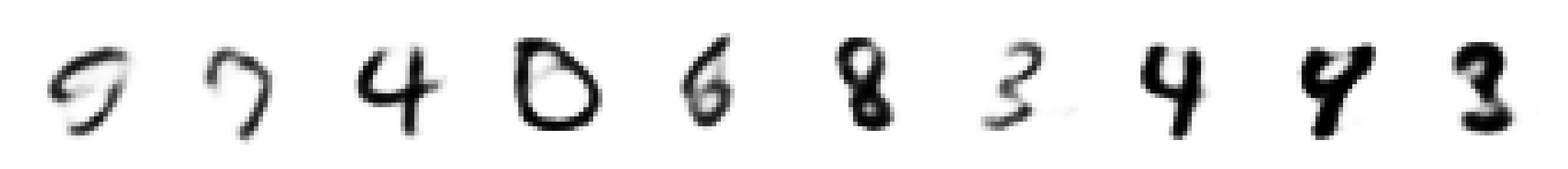}
		\end{subfigure}\\
		\begin{subfigure}[h]{0.22\linewidth}
			\raggedleft
			\tiny BILBO
		\end{subfigure}
		\hfil
		\begin{subfigure}[h]{0.75\linewidth}
			\includegraphics[width=\linewidth]{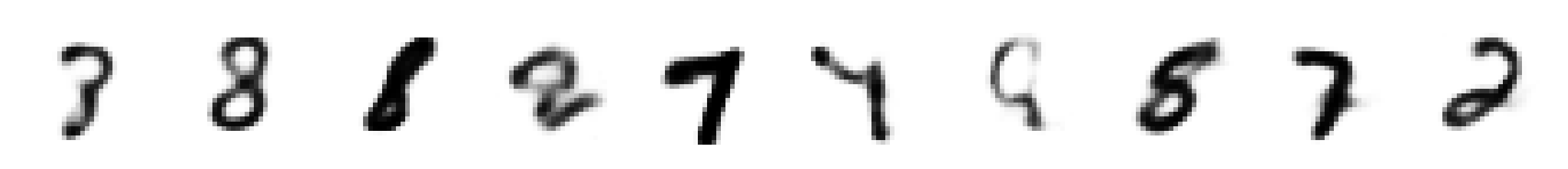}
		\end{subfigure}\\
		\caption{Generative samples.}\label{fig-generative-learned-vs-bilbo}
	\end{subfigure}	
	\caption{
		Experiments for MNIST with three methods for determining $\bm{\Sigma}$.
		``Learned $\bm{\Sigma}$'' \cref{eq-vae} with learned $\bm{\Sigma}$.
		``Constant $\bm{\Sigma}$'' proposed; \cref{eq-vae-s} with $\bm{\Sigma} = \mathbf{I}$ and $\mathbf{S}^2$ from \cref{eq-closed-s}.
		``BILBO'' proposed; \cref{eq-bilbo} with $\bm{\Sigma} = \mathbf{I}$ and $\mathbf{S}^2$ from \cref{eq-closed-s}.
		(a, b) progress during training (mean of five runs). The proposed methods produce a smaller variational lower bound than the learned variances method.
		(c) generative samples. Qualitatively, the three methods are indistinguishable.
	}\label{fig-elbo-burndown}
	\vspace{-0.2cm}
\end{figure}

\begin{figure}
	\hfil
	\begin{subfigure}[h]{0.25\linewidth}
		\includegraphics[width=\linewidth]{figures/scatter_mu_learned}
		\caption{Learned $\bm{\Sigma}$.}\label{fig-learned-bernoulli-scatter}
	\end{subfigure}
	\hfil
	\begin{subfigure}[h]{0.25\linewidth}
		\includegraphics[width=\linewidth]{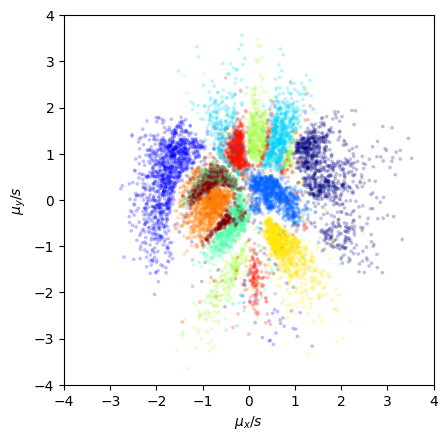}
		\caption{Constant $\bm{\Sigma}$.}\label{fig-constant-bernoulli-scatter}
	\end{subfigure}
	\hfil
	\begin{subfigure}[h]{0.25\linewidth}
		\includegraphics[width=\linewidth]{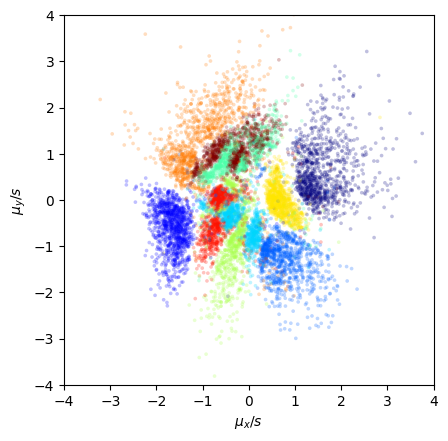}
		\caption{BILBO.}\label{fig-bilbo-bernoulli-scatter}
	\end{subfigure}
	\hfil
	\caption{
		Posterior means for a two-dimensional MNIST latent space colored by label, for three different methods for determining $\bm{\Sigma}$.
		(a) \cref{eq-vae} with learned $\bm{\Sigma}$.
		(b) proposed; \cref{eq-vae-s} with $\bm{\Sigma} = \mathbf{I}$ and $\mathbf{S}^2$ from \cref{eq-closed-s}.
		(c) proposed; \cref{eq-bilbo} with $\bm{\Sigma} = \mathbf{I}$ and $\mathbf{S}^2$ from \cref{eq-closed-s}.
		The proposed methods (b) and (c) produce a smaller ELBO than the traditional method (a) (\num{-134.2} and \num{-134.3} vs.\ \num{-136.5}).
	}\label{fig-bernoulli-scatter}
\end{figure}

We consider reformulating the expected variational lower bound as an expectation over a batch $B$. Recalling the optimal $\mathbf{S}^2$ in \cref{eq-closed-s}, substitution into \cref{eq-vae-s} with constant $\bm{\Sigma}$ yields:
\begin{align}\label{eq-reduced-elbo}
\e_{p(\mathbf{x})} [\text{ELBO}] = 
-\tfrac{1}{2} \log \det{\mathbf{I} + \bm{\Sigma}^\inv \e_{p(\mathbf{x})} [\bm{\mu}\bm{\mu}^\tran]}
+ \e_{p(\mathbf{x})} \! \left[\e_{\mathcal{N}(\mathbf{z}; \bm{\mu}, \bm{\Sigma})} [
\log p_{\bm{\theta}}(\mathbf{x}|\mathbf{z}) ]\right].
\end{align}
As we must estimate $\e_{p(\mathbf{x})} [\bm{\mu}\bm{\mu}^\tran]$ anyway (as $\bm{\Mu}_B$), this motivates us to define a simpler batch formulation, the \emph{Batch Information Lower Bound} (BILBO):
\begin{align}\label{eq-bilbo}
\text{BILBO} \equiv 
-\tfrac{1}{2} \log \det{\mathbf{I} + \bm{\Sigma}^\inv \bm{\Mu}_B}
+ \e_B [ \e_{\mathcal{N}(\mathbf{z}; \bm{\mu}, \bm{\Sigma})} [
\log p_{\bm{\theta}}(\mathbf{x}|\mathbf{z})
] ].
\end{align}
In practice, batch expectations are averages over batches. \cref{fig-elbo-burndown,fig-bernoulli-scatter} illustrate that the BILBO is as effective as the variational lower bound, both quantitatively and qualitatively, despite its simplicity.

\section{The Desolation of SMoG (Sampled Mixtures of Gaussians)}\label{sec-closed-likelihood}

The log likelihood term in \cref{eq-vae} is in general unbounded for learned likelihood variances.
This phenomenon has been noted in the literature \cite{pierrealex2018leveraging} and necessitates that VAE implementations impose artificial priors on the generative model, such as weight decay \cite{2013arXiv1312.6114K}, hand-tuned fixed variances \cite{DBLP:journals/corr/ZhaoSE17}, or bounded eigenvalues (effectively Tikhonov regularization) \cite{pierrealex2018leveraging}.
In this section we examine this problem in light of the exact log evidence from \cref{sec-noisy}, finding that any finite data set with Gaussian likelihood may be interpreted as a Sampled Mixture of Gaussians (SMoG) with zero KL divergence regardless of the data distribution, and that this immediately leads to the unboundedness of the likelihood if the likelihood variance is allowed to approach zero.

\paragraph{The True Posterior}
The left hand side of \cref{eq-elbo-diff} is equal to
$\mathrm{KL}(q_{\bm{\phi}}(\mathbf{z}|\mathbf{x}) | p_{\bm{\theta}}(\mathbf{z}|\mathbf{x}))$
by definition of the ELBO \cite{2013arXiv1312.6114K}. We also recognize the right hand side as the KL divergence of two Gaussian distributions having the same mean but different variances, which reveals the \emph{true posterior}:
\begin{align}
p_{\bm{\phi}}(\mathbf{z}|\mathbf{x}) = \mathcal{N}\left(\mathbf{z}; \bm{\mu}, (\mathbf{S}^\invsq + \mathbf{J}^\tran\!\mathbf{J})^\inv \right).
\label{eq-true-posterior}
\end{align}
At the stationary points of the ELBO, we find that the approximate posterior $q$ is equal to the true posterior $p$, with zero KL divergence. But this ought to be intractable, so what's going on?
The answer is that we do not have an infinite data source; we have a \emph{sampled} data source. Computing expectations over an infinite data source may be intractable, but expectation over a finite number of data samples leads to a ``true posterior'' that is a Mixture of Gaussians (MoG). No matter how many data samples we use, we may always interpret the data as a sampled mixture of likelihood functions, in our case a Sampled Mixture of Gaussians (SMoG), and simply use enough Gaussians to obtain zero KL divergence. This defect is of more philosophical than practical importance, but it does introduce a problem with learned likelihoods which we explore next.

\begin{figure}[ht]
\begin{subfigure}[h]{\linewidth}
\begin{subfigure}[h]{0.32\linewidth}
	\begin{subfigure}[h]{0.22\linewidth}
	\tiny \quad
	\end{subfigure}
	\begin{subfigure}[h]{0.75\linewidth}
		\centering
		\tiny 2 latent dimensions
	\end{subfigure}\\
	\begin{subfigure}[h]{0.22\linewidth}
		\raggedleft
		\tiny Learned $\bm{\Tau}$
	\end{subfigure}
	\hfil
	\begin{subfigure}[h]{0.75\linewidth}
		\includegraphics[width=\linewidth]{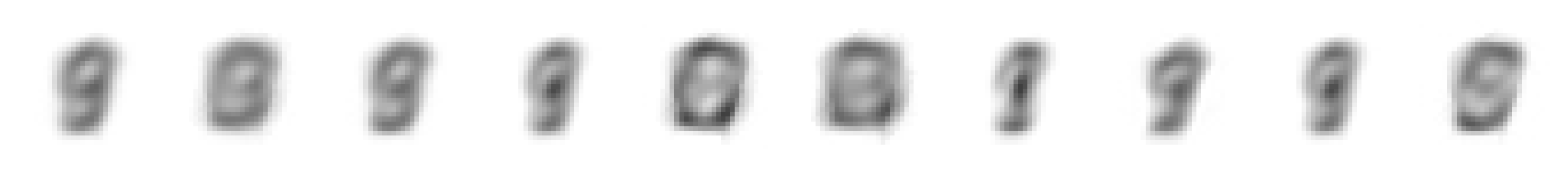}
	\end{subfigure}\\
	\begin{subfigure}[h]{0.22\linewidth}
		\raggedleft
		\tiny $\lambda\!=\!\tfrac{3}{4}$
	\end{subfigure}
	\hfil
	\begin{subfigure}[h]{0.75\linewidth}
		\includegraphics[width=\linewidth]{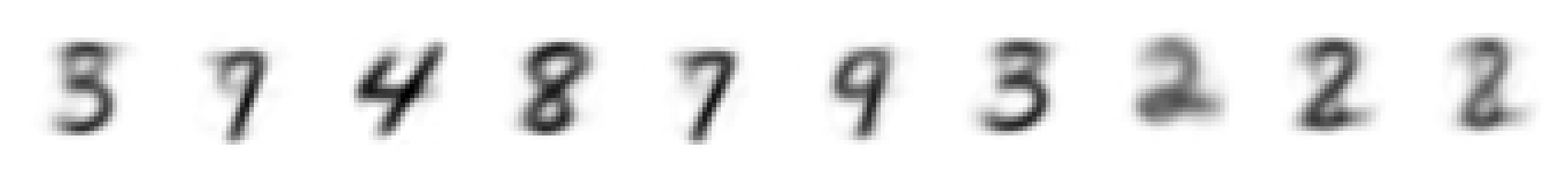}
	\end{subfigure}\\
	\begin{subfigure}[h]{0.22\linewidth}
		\raggedleft
		\tiny $\lambda\!=\!1$
	\end{subfigure}
	\hfil
	\begin{subfigure}[h]{0.75\linewidth}
		\includegraphics[width=\linewidth]{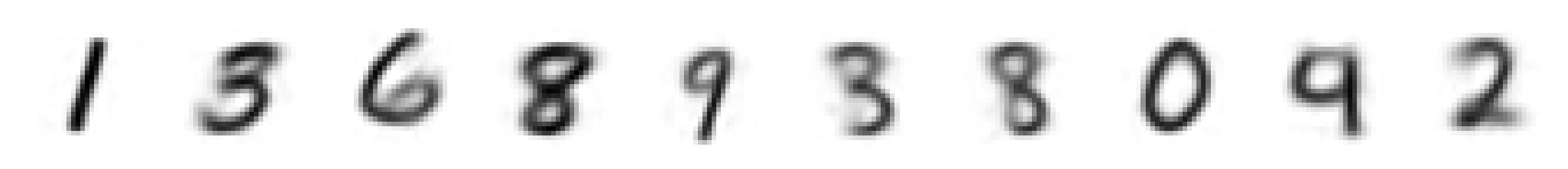}
	\end{subfigure}\\
	\begin{subfigure}[h]{0.22\linewidth}
		\raggedleft
		\tiny $\lambda\!=\!2$
	\end{subfigure}
	\hfil
	\begin{subfigure}[h]{0.75\linewidth}
		\includegraphics[width=\linewidth]{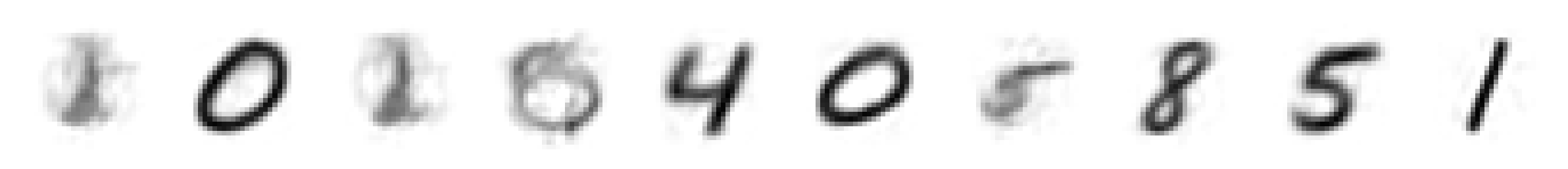}
	\end{subfigure}\\
	\begin{subfigure}[h]{0.22\linewidth}
		\raggedleft
		\tiny $\lambda\!=\!5$
	\end{subfigure}
	\hfil
	\begin{subfigure}[h]{0.75\linewidth}
		\includegraphics[width=\linewidth]{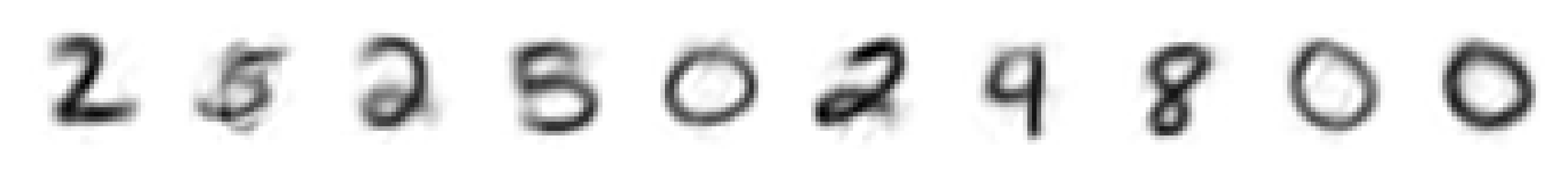}
	\end{subfigure}\\
	\begin{subfigure}[h]{0.22\linewidth}
		\raggedleft
		\tiny $\lambda\!=\!10$
	\end{subfigure}
	\hfil
	\begin{subfigure}[h]{0.75\linewidth}
		\includegraphics[width=\linewidth]{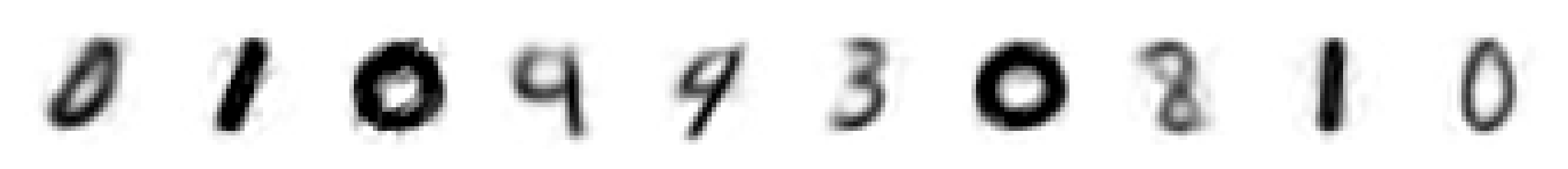}
	\end{subfigure}\\
\end{subfigure}
\hfil
\begin{subfigure}[h]{0.32\linewidth}
	\begin{subfigure}[h]{0.22\linewidth}
		\tiny \quad
	\end{subfigure}
	\begin{subfigure}[h]{0.75\linewidth}
		\centering
		\tiny 2 latent dimensions
	\end{subfigure}\\
	\begin{subfigure}[h]{0.22\linewidth}
		\raggedleft
		\tiny $\lambda\!=\!\nicefrac{1}{2}$
	\end{subfigure}
	\hfil
	\begin{subfigure}[h]{0.75\linewidth}
		\includegraphics[width=\linewidth]{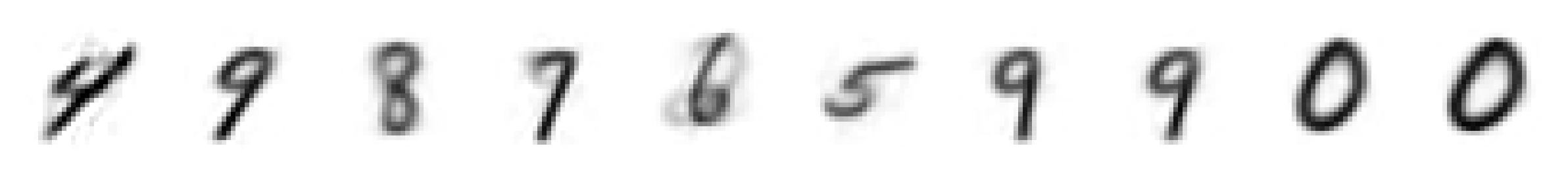}
	\end{subfigure}\\
	\begin{subfigure}[h]{0.22\linewidth}
		\raggedleft
		\tiny $\lambda\!=\!1$
	\end{subfigure}
	\hfil
	\begin{subfigure}[h]{0.75\linewidth}
		\includegraphics[width=\linewidth]{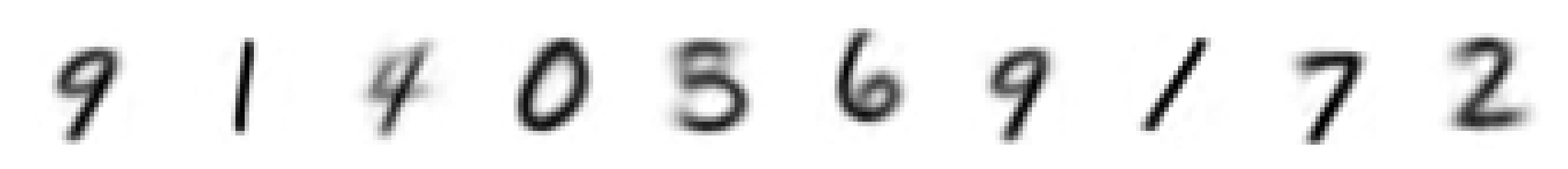}
	\end{subfigure}\\
	\begin{subfigure}[h]{0.22\linewidth}
		\raggedleft
		\tiny $\lambda\!=\!10$
	\end{subfigure}
	\hfil
	\begin{subfigure}[h]{0.75\linewidth}
		\includegraphics[width=\linewidth]{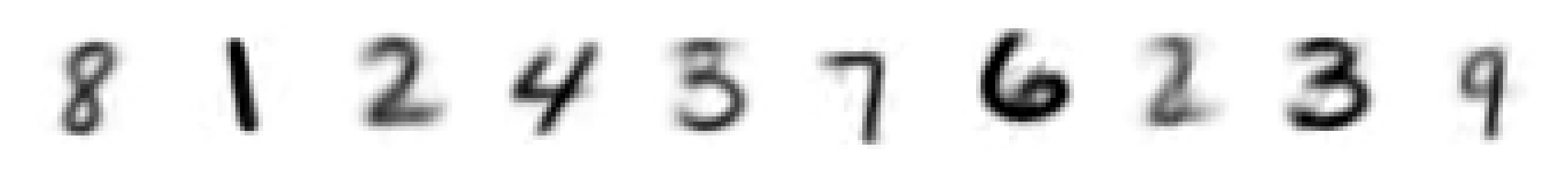}
	\end{subfigure}\\
	\begin{subfigure}[h]{0.22\linewidth}
		\raggedleft
		\tiny $\lambda\!=\!100$
	\end{subfigure}
	\hfil
	\begin{subfigure}[h]{0.75\linewidth}
		\includegraphics[width=\linewidth]{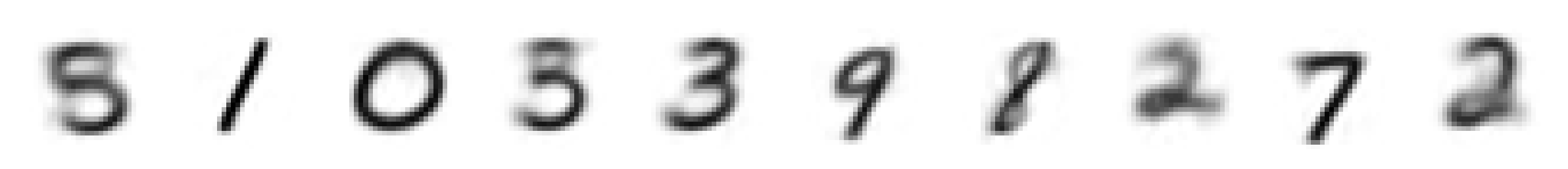}
	\end{subfigure}\\
\end{subfigure}
\hfil
\begin{subfigure}[h]{0.32\linewidth}
	\begin{subfigure}[h]{0.22\linewidth}
		\tiny \quad
	\end{subfigure}
	\begin{subfigure}[h]{0.75\linewidth}
		\centering
		\tiny 2 latent dimensions
	\end{subfigure}\\
	\begin{subfigure}[h]{0.22\linewidth}
		\raggedleft
		\tiny $\tau\!=\!\nicefrac{1}{10}$
	\end{subfigure}
	\hfil
	\begin{subfigure}[h]{0.75\linewidth}
		\includegraphics[width=\linewidth]{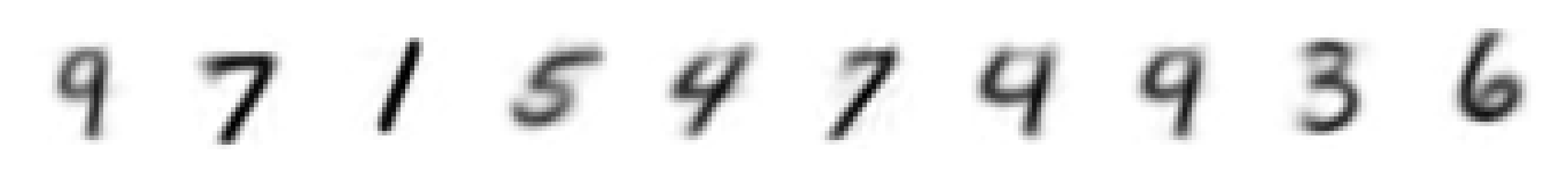}
	\end{subfigure}\\
	\begin{subfigure}[h]{0.22\linewidth}
		\raggedleft
		\tiny $\tau\!=\!\nicefrac{1}{5}$
	\end{subfigure}
	\hfil
	\begin{subfigure}[h]{0.75\linewidth}
		\includegraphics[width=\linewidth]{figures/sample_baggins_t02}
	\end{subfigure}\\
	\begin{subfigure}[h]{0.22\linewidth}
		\raggedleft
		\tiny $\tau\!=\!\nicefrac{1}{2}$
	\end{subfigure}
	\hfil
	\begin{subfigure}[h]{0.75\linewidth}
		\includegraphics[width=\linewidth]{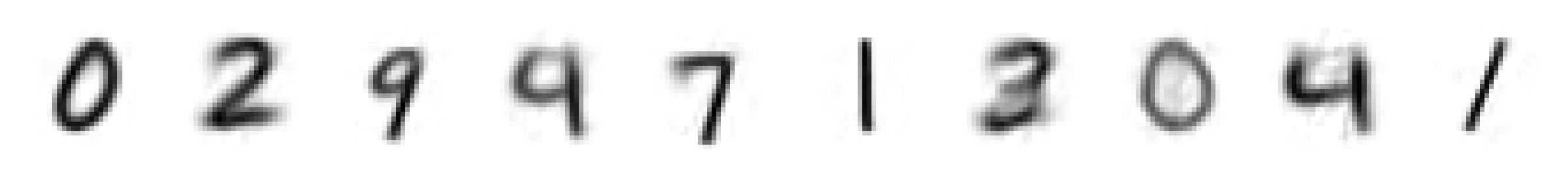}
	\end{subfigure}\\
	\begin{subfigure}[h]{0.22\linewidth}
		\raggedleft
		\tiny $\tau\!=\!1$
	\end{subfigure}
	\hfil
	\begin{subfigure}[h]{0.75\linewidth}
		\includegraphics[width=\linewidth]{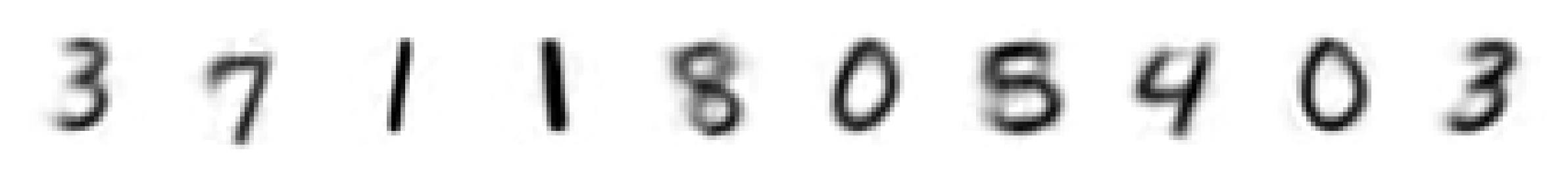}
	\end{subfigure}\\
	\begin{subfigure}[h]{0.22\linewidth}
		\raggedleft
		\tiny $\tau\!=\!5$
	\end{subfigure}
	\hfil
	\begin{subfigure}[h]{0.75\linewidth}
		\includegraphics[width=\linewidth]{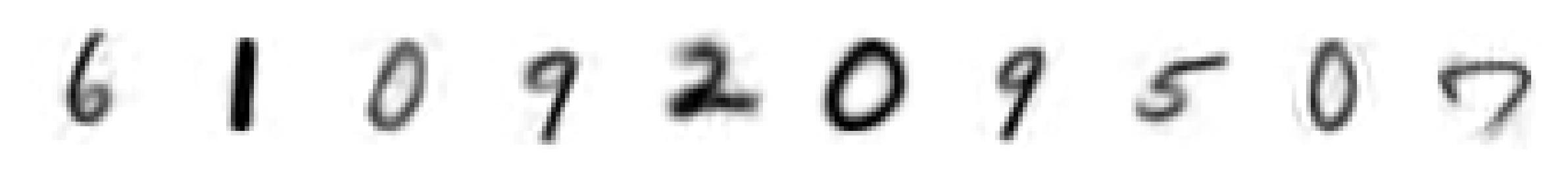}
	\end{subfigure}\\
\end{subfigure}
\end{subfigure}
\begin{subfigure}[h]{\linewidth}
	\begin{subfigure}[h]{0.32\linewidth}
		\begin{subfigure}[h]{0.22\linewidth}
			\tiny \quad
		\end{subfigure}
		\begin{subfigure}[h]{0.75\linewidth}
			\centering
			\tiny 100 latent dimensions
		\end{subfigure}\\
		\begin{subfigure}[h]{0.22\linewidth}
			\raggedleft
			\tiny Learned $\bm{\Tau}$
		\end{subfigure}
		\hfil
		\begin{subfigure}[h]{0.75\linewidth}
			\includegraphics[width=\linewidth]{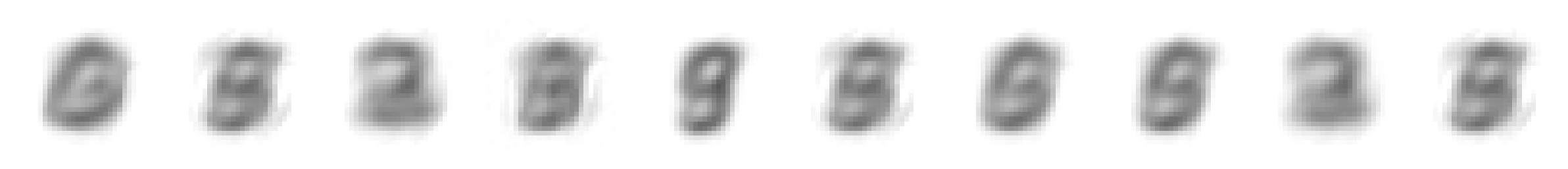}
		\end{subfigure}\\
		\begin{subfigure}[h]{0.22\linewidth}
			\raggedleft
			\tiny $\lambda\!=\!\tfrac{3}{4}$
		\end{subfigure}
		\hfil
		\begin{subfigure}[h]{0.75\linewidth}
			\includegraphics[width=\linewidth]{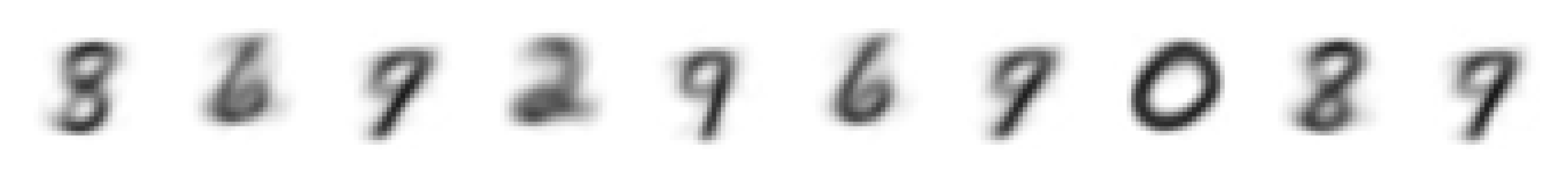}
		\end{subfigure}\\
		\begin{subfigure}[h]{0.22\linewidth}
			\raggedleft
			\tiny $\lambda\!=\!1$
		\end{subfigure}
		\hfil
		\begin{subfigure}[h]{0.75\linewidth}
			\includegraphics[width=\linewidth]{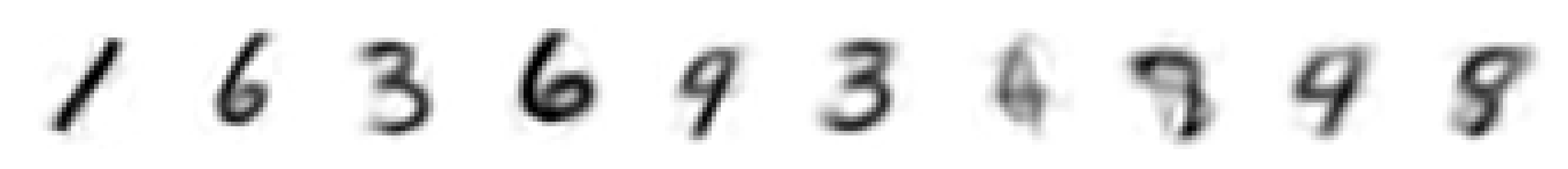}
		\end{subfigure}\\
		\begin{subfigure}[h]{0.22\linewidth}
			\raggedleft
			\tiny $\lambda\!=\!2$
		\end{subfigure}
		\hfil
		\begin{subfigure}[h]{0.75\linewidth}
			\includegraphics[width=\linewidth]{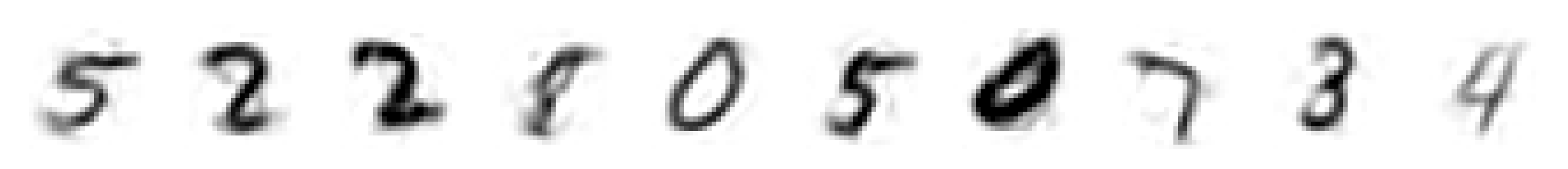}
		\end{subfigure}\\
		\begin{subfigure}[h]{0.22\linewidth}
			\raggedleft
			\tiny $\lambda\!=\!5$
		\end{subfigure}
		\hfil
		\begin{subfigure}[h]{0.75\linewidth}
			\includegraphics[width=\linewidth]{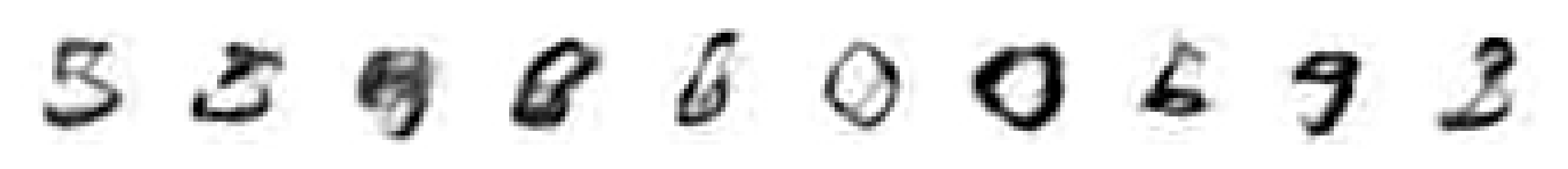}
		\end{subfigure}\\
		\begin{subfigure}[h]{0.22\linewidth}
			\raggedleft
			\tiny $\lambda\!=\!10$
		\end{subfigure}
		\hfil
		\begin{subfigure}[h]{0.75\linewidth}
			\includegraphics[width=\linewidth]{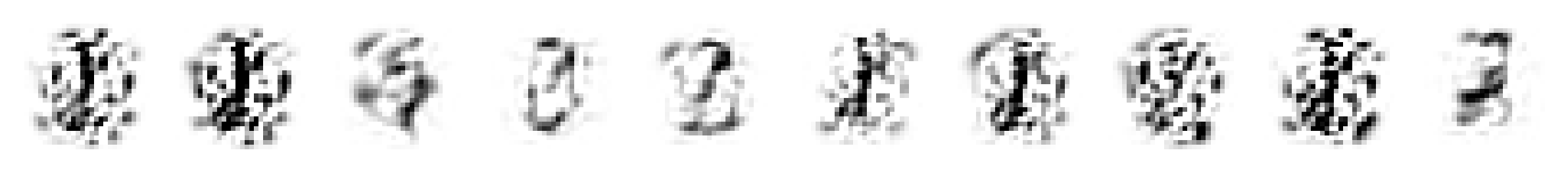}
		\end{subfigure}\\
		\vspace{0.1ex}
	\end{subfigure}
	\hfil
	\begin{subfigure}[h]{0.32\linewidth}
		\begin{subfigure}[h]{0.22\linewidth}
			\tiny \quad
		\end{subfigure}
		\begin{subfigure}[h]{0.75\linewidth}
			\centering
			\tiny 100 latent dimensions
		\end{subfigure}\\
		\begin{subfigure}[h]{0.22\linewidth}
			\raggedleft
			\tiny $\lambda\!=\!\nicefrac{1}{2}$
		\end{subfigure}
		\hfil
		\begin{subfigure}[h]{0.75\linewidth}
			\includegraphics[width=\linewidth]{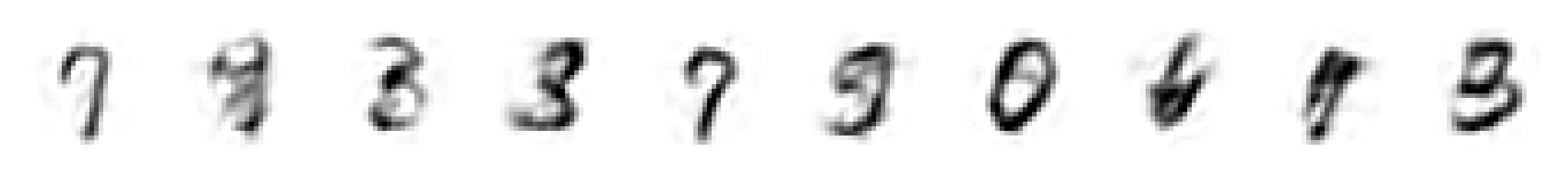}
		\end{subfigure}\\
		\begin{subfigure}[h]{0.22\linewidth}
			\raggedleft
			\tiny $\lambda\!=\!1$
		\end{subfigure}
		\hfil
		\begin{subfigure}[h]{0.75\linewidth}
			\includegraphics[width=\linewidth]{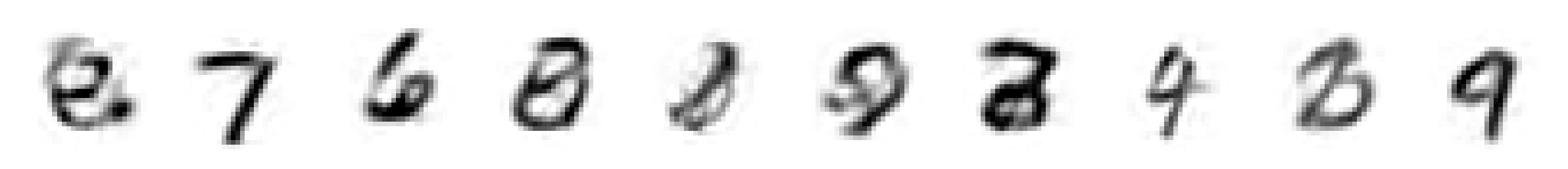}
		\end{subfigure}\\
		\begin{subfigure}[h]{0.22\linewidth}
			\raggedleft
			\tiny $\lambda\!=\!10$
		\end{subfigure}
		\hfil
		\begin{subfigure}[h]{0.75\linewidth}
			\includegraphics[width=\linewidth]{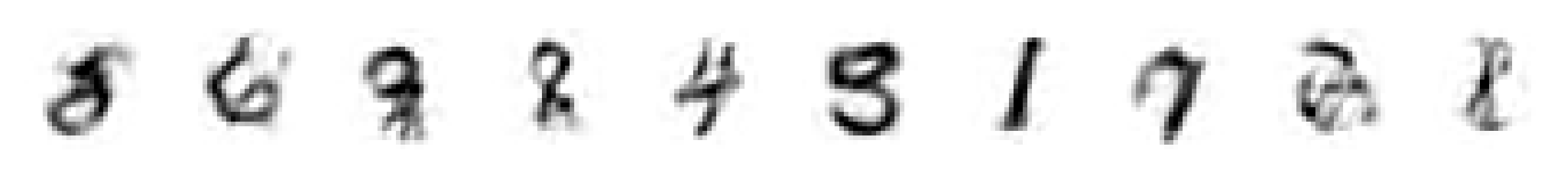}
		\end{subfigure}\\
		\begin{subfigure}[h]{0.22\linewidth}
			\raggedleft
			\tiny $\lambda\!=\!100$
		\end{subfigure}
		\hfil
		\begin{subfigure}[h]{0.75\linewidth}
			\includegraphics[width=\linewidth]{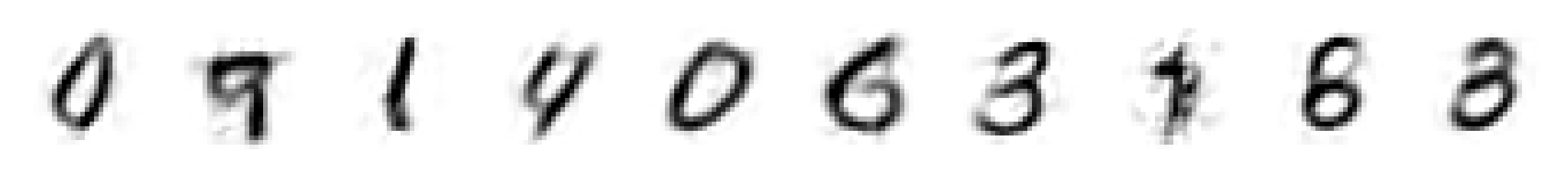}
		\end{subfigure}\\
	\end{subfigure}
	\hfil
	\begin{subfigure}[h]{0.32\linewidth}
		\begin{subfigure}[h]{0.22\linewidth}
			\tiny \quad
		\end{subfigure}
		\begin{subfigure}[h]{0.75\linewidth}
			\centering
			\tiny 100 latent dimensions
		\end{subfigure}\\
		\begin{subfigure}[h]{0.22\linewidth}
			\raggedleft
			\tiny $\tau\!=\!\nicefrac{1}{10}$
		\end{subfigure}
		\hfil
		\begin{subfigure}[h]{0.75\linewidth}
			\includegraphics[width=\linewidth]{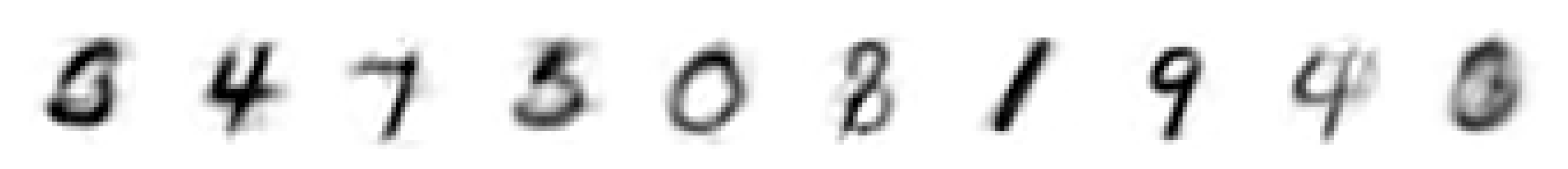}
		\end{subfigure}\\
		\begin{subfigure}[h]{0.22\linewidth}
			\raggedleft
			\tiny $\tau\!=\!\nicefrac{1}{5}$
		\end{subfigure}
		\hfil
		\begin{subfigure}[h]{0.75\linewidth}
			\includegraphics[width=\linewidth]{figures/sample_baggins100_t02}
		\end{subfigure}\\
		\begin{subfigure}[h]{0.22\linewidth}
			\raggedleft
			\tiny $\tau\!=\!\nicefrac{1}{2}$
		\end{subfigure}
		\hfil
		\begin{subfigure}[h]{0.75\linewidth}
			\includegraphics[width=\linewidth]{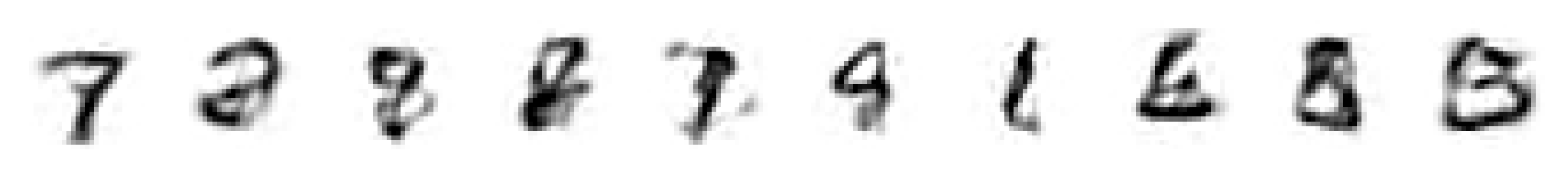}
		\end{subfigure}\\
		\begin{subfigure}[h]{0.22\linewidth}
			\raggedleft
			\tiny $\tau\!=\!1$
		\end{subfigure}
		\hfil
		\begin{subfigure}[h]{0.75\linewidth}
			\includegraphics[width=\linewidth]{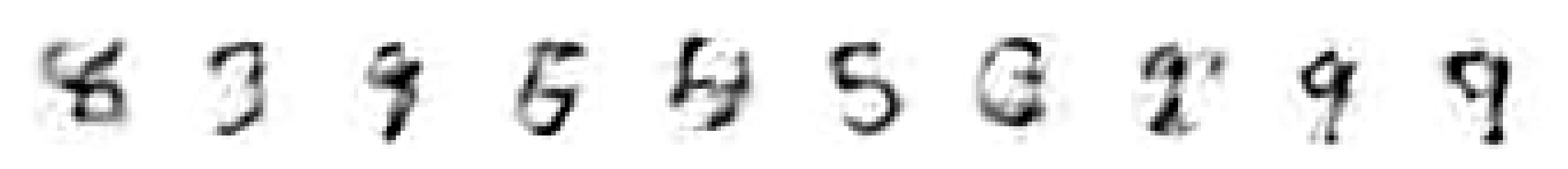}
		\end{subfigure}\\
		\begin{subfigure}[h]{0.22\linewidth}
			\raggedleft
			\tiny $\tau\!=\!5$
		\end{subfigure}
		\hfil
		\begin{subfigure}[h]{0.75\linewidth}
			\includegraphics[width=\linewidth]{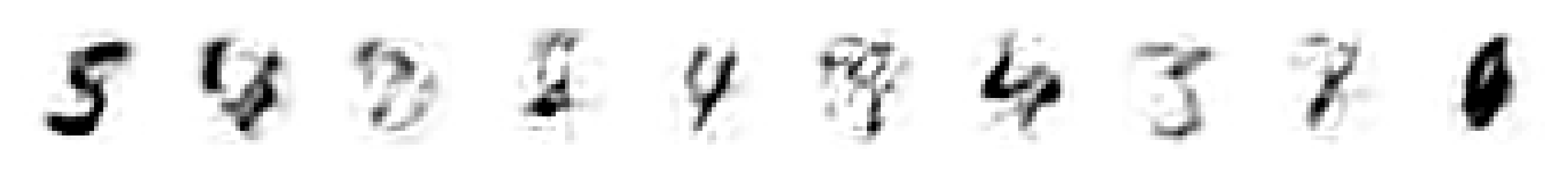}
		\end{subfigure}\\
	\end{subfigure}
\end{subfigure}
\begin{subfigure}[h]{0.32\linewidth}
\caption{Learned and constant $\bm{\Tau}$.}\label{fig-unbounded-tau}
\end{subfigure}
\hfil
\begin{subfigure}[h]{0.32\linewidth}
	\caption{BAGGINS; varying $\lambda$.}\label{fig-generative-baggins-consistency}
\end{subfigure}
\hfil
\begin{subfigure}[h]{0.32\linewidth}
	\caption{BAGGINS; varying $\tau$.}\label{fig-generative-baggins-tau}
\end{subfigure}
\caption{Generative samples for MNIST with Gaussian likelihood. $\lambda$ indicates a scaling factor applied to the data vectors. 
(a) learned $\bm{\Tau}$ fails to converge without further constraint, while constant $\bm{\Tau}=\mathbf{I}$ is extremely sensitive to the data scale.
(b) BAGGINS with $\tau\!=\!\nicefrac{1}{5}$ (\cref{eq-baggins}) and varying $\lambda$. (Contrast to (a).) Qualitatively, the samples are stable regardless of the data scale.
(c) BAGGINS with varying $\tau$, the information factor, which affects the sample quality.
}
\vspace{-0.15cm}
\end{figure}

\paragraph{The Optimal Likelihood Variance is Zero}
Recall that the claims in \cref{sec-noisy} extend to any Riemannian likelihood function, in particular that the ELBO is at its stationary points equal to log evidence for noisy data, where the noise model is induced by the likelihood function. In the limit of zero likelihood variance, it reduces to the noiseless log evidence. Any increase in the likelihood variance is equivalent to blurring the estimated data density function, which strictly increases entropy. As entropy is the negative expected log evidence, equal to the negative expected ELBO at its stationary points, any increase in the likelihood variance strictly decreases the maximum ELBO. Therefore the maximum ELBO is obtained at zero likelihood variance.
For example consider a Gaussian likelihood with $p_{\bm{\theta}}(\mathbf{x}|\mathbf{z}) = \mathcal{N}(\mathbf{x}; \bm{\nu_\theta}( \mathbf{z}), \bm{\Tau}_{\bm{\theta}}( \mathbf{z}))$. Dropping the subscripts and writing $\bm{\Tau} \equiv \bm{\Tau}_{\bm{\theta}}(\bm{\mu}_{\bm{\phi}}(\mathbf{x}))$, the log evidence as in \cref{eq-exact-noisy} but with $\bm{\gamma} = \bm{\Tau}^\invsqrt (\mathbf{x} - \bm{\nu}(\bm{\mu}))$ and $\mathbf{x} = \bm{\nu}(\mathbf{S}\bm{\zeta}) + \sqrt{\bm{\Tau}} \bm{\gamma}$ is:
\begin{align}
\log p_{\bm{\theta}}(\mathbf{x})
=
- \tfrac{1}{2} \big(
m \log 2 \pi
+ |\mathbf{S}^\inv\bm{\mu}|^2
+ (\mathbf{x} - \bm{\nu}(\bm{\mu}))^\tran
\bm{\Tau}^\inv
(\mathbf{x} - \bm{\nu}(\bm{\mu}))
+ \log \det{\mathbf{J} \mathbf{S}^2\! \mathbf{J}^\tran \!+\! \bm{\Tau}}
\big).
\label{eq-exact-tau}
\end{align}
At the limit $\bm{\Tau} \rightarrow \mathbf{0}$, the data is entirely modeled by the latent process (assuming $n = m$ for now):
\begin{align}
\lim_{\bm{\Tau} \rightarrow \mathbf{0}}
\log p_{\bm{\theta}}(\mathbf{x})
=
- \tfrac{1}{2} \big(
m \log 2 \pi
+ |\mathbf{S}^\inv\bm{\mu}|^2
+ \log \det{\mathbf{J} \mathbf{S}^2\! \mathbf{J}^\tran}
\big).
\label{eq-exact-tau-limit}
\end{align}
%
As we may let $\mathbf{S} = \mathbf{I}$ without affecting the ELBO, we recognize \cref{eq-exact-tau-limit} as the log likelihood by a change of variables, equivalent to the objective employed in normalizing flow networks \cite{dinh2017density}. 
There is, however, a problem. When $\bm{\Tau}$ is small, the data density is modeled as a tiny island of data density at each sample in the Sampled Mixture of Gaussians. The islands are essentially independent and hence the decoder is free to shrink-wrap them individually, leading to $\mathbf{J} \mathbf{S}^2\! \mathbf{J}^\tran \rightarrow \mathbf{0}$ as $\bm{\Tau} \rightarrow \mathbf{0}$.
While some families of likelihood models naturally constrain the variance such as the Bernoulli distribution, others such as our Gaussian likelihood example cannot converge without further constraint.
\cref{fig-unbounded-tau} illustrates the failure of directly learning $\bm{\Tau}$, and the sensitivity of hand-tuned likelihood variances to the intrinsic scale of the data, producing blurry generative samples when the variance is too high, and unrecognizable generative samples when the variance is too low.

\subsection{Bounded Aggregate Information Sampling}\label{sec-baggins}

To address the instability of learned likelihood variances and the difficulty in hand tuning them, we propose a novel likelihood constraint that is invariant to scale, based on both the variance and the information content of the data population, with one free parameter $\tau$:
\begin{align}
    \tau (\mathbf{x} - \bm{\nu}(\bm{\mu}))^\tran
        \bm{\Tau}^\inv
        (\mathbf{x} - \bm{\nu}(\bm{\mu}))
    = \tr \bm{\Sigma}^\inv \mathbf{S}^2.
    \label{eq-baggins-constraint}
\end{align}
The left hand side estimates the variance of the data noise process (scaled by $\tau$) and the right hand side is a proxy to the information, which increases monotonically with the log determinant term in \cref{eq-exact-noisy}. This formulation may be rearranged to a single sample estimator assuming isotropic $\bm{\Tau}$:
%
\begin{align}
    \frac{\tau |\mathbf{x} - \bm{\nu}(\mathbf{z})|^2}
    {\tr \bm{\Sigma}^\inv \mathbf{S}^2}
    \mathbf{I}_m
    \estimates
    \bm{\Tau}.
    \label{eq-baggins}
\end{align}
In a deep network setting, this allows us to aggregate the bounded observed information (negative log likelihood) over a batch, which we call \emph{Bounded Aggregate Information Sampling} (BAGGINS). We form a scaling matrix $\mathbf{S}^2_B$ for batch $B$ via \cref{eq-closed-s} to compute $\bm{\Tau}$ using \cref{eq-baggins}. This may be plugged into \cref{eq-vae-s} as usual, or combined with \cref{eq-bilbo} to yield:
\begin{equation}\label{eq-bilbo-baggins}
\text{BILBO}_{\text{BAGGINS}}
\equiv 
-\tfrac{1}{2}\! \left(
    \log \det{\mathbf{I} + \bm{\Sigma}^\inv \bm{\Mu}_B}
    + \tfrac{1}{\tau} \tr (\mathbf{I} + \bm{\Sigma}^\inv \bm{\Mu}_B)
    + \e_B [ \log \det{ 2\pi \bm{\Tau} } ] \right),
\end{equation}
with $\mathbf{z} \sim \mathcal{N}(\bm{\mu}, \bm{\Sigma})$ independently for each example in the batch. In a deep network setting, this eliminates half of the outputs of the likelihood model. \cref{fig-generative-baggins-consistency} demonstrates invariance to data scaling, with generative samples comparable to hand-tuned likelihoods. \cref{fig-generative-baggins-tau} illustrates the effect that different values of the information factor $\tau$ has on the quality of the generative samples.
While many methods exist to regularize the likelihood variances, ours directly targets the underlying stability problem while producing a desirable characteristic (scale invariance). We hope that our exposition may lead to more intuitive understanding of VAE likelihoods.

\section{Discussion and Implications}\label{sec-discussion}

\paragraph{These Are Not Update Rules}
A possible point of confusion is the form of the equations in this work relating various parameters, which might be mistaken for update rules in an iterative optimization approach. We wish to clarify our intent: the equations hold at stationary points of the ELBO, and are employed in the derivation of our results. The actual implementations in our experiments follow ordinary practices in neural network training, with only minor changes to the problem formulation (e.g.\ constant posterior variances rather than learned, and estimated priors rather than given).

\paragraph{On Network Capacity}
As our results are derived from the ELBO without consideration for any particular network architecture, one may wonder if our results still hold for practical implementations of networks with limited capacity. For example, we prove that the ideal optimal decoder is smooth, and the ideal optimal ELBO is a tight bound on the log evidence, but what about networks that lack the capacity to represent the ideal decoder exactly?
The answer lies in the universal approximation property of neural networks \cite{HORNIK1989359,pmlr-v125-kidger20a}. Training a neural network will pursue an approximation to the ideal result, regardless of the capacity of the network. We make no statement about the smoothness of the neural network itself, nor the smoothness of the search landscape during training, nor the smoothness of the training gradients. Our results rely only on the optimal theoretical decoder.

\paragraph{On Memorization}
One might wonder if the closed form decoder in \cref{eq-closed-warp} implies that the VAE is simply memorizing the data because it depends on every data example. In fact it does not. Rather, the encoder removes one standard deviation of information according to the noise model induced by the likelihood function.
To understand this, consider the maximization of the expected value of \cref{eq-exact-noisy} when the data distribution is anisotropic Gaussian, with $n=m$. Without loss of generality, we may let $\mathbf{S} = \mathbf{I}$ and orient the latent space so its cardinal axes align with the principal axes of the data variance. In this orientation, \cref{eq-exact-noisy} is separable and maximizing the expectation over each dimension of the Gaussian is tractable. We find that along any dimension where the data variance is not greater than the likelihood variance, the corresponding latent dimension collapses to $0$, as does the corresponding column of the decoder Jacobian (or $\mathbf{J}$), leaving only the residual term. In other words, any subspace of the data with less variance than the likelihood is modeled as noise, and zeroed out during encoding. Other subspaces are reduced but not zeroed out. There is no hope to recover the information in the lost dimensions despite the relationship between the residual and the latent vector in \cref{eq-closed-mu}, because the zero columns in $\mathbf{J}$ are not invertible. Since the decoder varies smoothly, this argument extends to non-Gaussian data by interpreting the data distribution as Gaussian with spatially varying parameters, where in general subspaces of $\bm{\mu}$ and $\mathbf{J}$ will be zeroed out rather than rows and columns.
Moreover, by the same assumptions that motivate VAEs, we expect the intrinsic data manifold to have low rank relative to the ambient space, and hence we expect $\mathbf{J}$ to be of low rank (i.e.\ most eigenvalues zeroed out). Therefore the majority of the information is forgotten during encoding, and cannot be remembered in decoding.
Memorization only occurs when the data examples are distant from their neighbors relative to the standard deviation of the likelihood, indicating a poor likelihood function or insufficient data, which are problems common to all VAEs.

\paragraph{On Blurriness}
None of the optimality conditions of the ELBO in \cref{sec-solutions} demand a reduction in the spatial sharpness of reconstructed data, so we may state that VAE reconstructions are not inherently blurry. However, VAEs do reduce \emph{variation across examples}, for example by the reductions in the encoder mentioned in the previous paragraph, or by dimension reduction ($n < m$). The relationship between variation reduction and sharpness is determined solely by the likelihood function (or loss function in deep learning parlance). If there are any blurry objects along the geodesic interpolating two sharp examples in the Riemannian space induced by the likelihood function, then reducing variation may produce those blurry objects. For example, the sum-of-squared-residuals loss induces linear interpolations resulting in ghosting and thereby blurriness. The remedy is to choose loss functions for which the lowest cost path between two sharp examples contains only sharp objects.

\section{Experiments}\label{sec-experiments}

We perform experiments on deep networks using TensorFlow \cite{tensorflow2015-whitepaper}. We use only simple network architectures since our goal is to test our theoretical results rather than a state-of-the-art algorithm.
For the MNIST dataset \cite{lecun-mnisthandwrittendigit-2010}, the encoder and decoder are similar, with four densely connected layers having \num{200} hidden units, and ReLU activations in the middle layers. The encoder output is interpreted as logits for Bernoulli likelihood models and as means for Gaussian likelihood models. When learned variances are employed (in either the encoder or decoder), an additional output layer is included with SoftPlus activations, and interpreted as standard deviations.
Training is performed using the training set of \num{60000} examples and all reported lower bound estimates and scatter plots are produced using the test set of \num{10000} examples.
We run experiments with \num{2} dimensional and \num{100} dimensional latent spaces using the Adam Optimizer \cite{Adam} with a learning rate of \num{0.001} and batch size of \num{300} for \num{200} epochs, except for the \num{100} dimensional Gaussian likelihood experiments which we ran for \num{400} epochs.
For the CelebA dataset \cite{liu2015faceattributes} we use the network architecture from \cite{DBLP:journals/corr/Barron17} with the RGB Pixel model and \num{16} latent dimensions, trained for \num{10000} iterations.

\subsection{Experiments with Posterior Variances}

\begin{figure}[ht]
	\centering
	\begin{subfigure}[h]{0.35\linewidth}
		\begin{subfigure}[h]{0.1\linewidth}\tiny\caption{}\end{subfigure}
		\begin{subfigure}[h]{0.88\linewidth}
			\includegraphics[width=\linewidth]{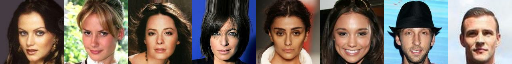}
		\end{subfigure}\\
		\begin{subfigure}[h]{0.1\linewidth}\caption{}\end{subfigure}
		\begin{subfigure}[h]{0.88\linewidth}
			\includegraphics[width=\linewidth]{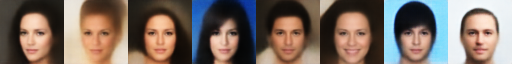}
		\end{subfigure}\\
		\begin{subfigure}[h]{0.1\linewidth}\caption{}\end{subfigure}
		\begin{subfigure}[h]{0.88\linewidth}
			\includegraphics[width=\linewidth]{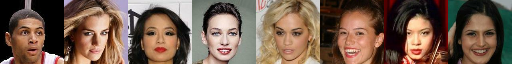}
		\end{subfigure}\\
		\begin{subfigure}[h]{0.1\linewidth}\caption{}\end{subfigure}
		\begin{subfigure}[h]{0.88\linewidth}
			\includegraphics[width=\linewidth]{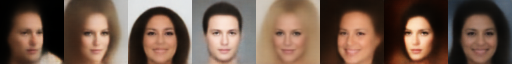}
		\end{subfigure}\\
		\begin{subfigure}[h]{0.98\linewidth}\caption*{Reconstructions.}\end{subfigure}
	\end{subfigure}
	\begin{subfigure}[h]{0.31\linewidth}
		\vspace*{-0.5ex}\includegraphics[width=\linewidth]{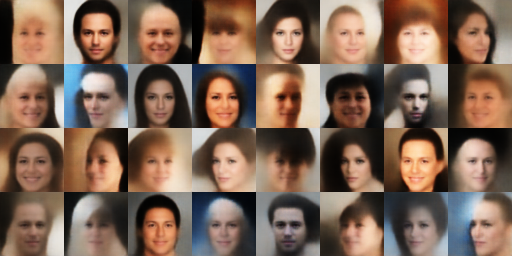}%
		\vspace*{-0.5ex}\caption{Samples using ELBO.}
	\end{subfigure}
	\begin{subfigure}[h]{0.31\linewidth}
		\vspace*{-0.5ex}\includegraphics[width=\linewidth]{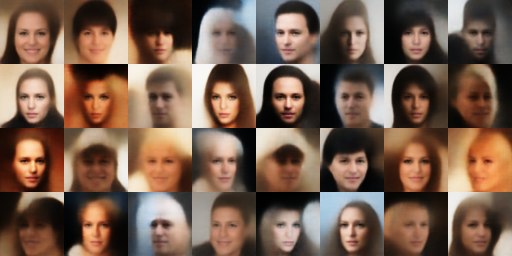}%
		\vspace*{-0.5ex}\caption{Samples using BILBO.}
	\end{subfigure}%
	\vspace{-1em}
	\caption{
		(a, c) input data for reconstructions using a VAE trained with (b) standard practice using ELBO and (d) BILBO. 
		(e) generative samples for CelebA using Standard VAE with ELBO loss but sampling from the optimal prior from \cref{eq-closed-s}, and
		(f) using BILBO loss with constant posterior variance $\bm{\Sigma} = \mathbf{I}/100$.
		The two methods produce qualitatively similar results.
	}
	\vspace{-1em}
	\label{fig-celeb}
\end{figure}

We trained a standard VAE with learned posterior variances and Bernoulli likelihood using \cref{eq-vae}. We also trained a VAE with constant posterior variance using \cref{eq-vae-s}, and another using the BILBO objective (\cref{eq-bilbo}), both using the prior variance from \cref{eq-closed-s}. \cref{fig-elbo-burndown} compares the three methods. The learned variances method lags significantly behind the two proposed methods, validating the proposed reformulations.
\cref{fig-generative-learned-vs-bilbo} shows little qualitative difference between generative samples from the three methods,
and \cref{fig-bernoulli-scatter} shows qualitatively similar distributions over the latent space.
(See \cref{fig-bilbo-scatter-sigma,fig-learned-manifold,fig-bilbo-manifold} for additional results.)
We performed further experiments on the more complex CelebA dataset. \cref{fig-celeb} illustrates results comparing the standard ELBO objective vs.\ the BILBO. Both methods yielded comparable ELBO and loss. We show generative samples from the optimal prior (\cref{eq-closed-s}) for both methods, since sampling the standard normal prior exhibited qualitatively less variation than the reconstructions. We found that choosing $\bm{\Sigma} = \mathbf{I}$ resulted in large optimal prior variances ($\tr\mathbf{S}^2 \sim$ \num{25000}) which seemed to slightly affect the quality of generative samples. We attribute this to the complexity of the dataset, hence choosing a smaller $\bm{\Sigma} = \mathbf{I}/100$ is more appropriate despite mathematically equivalent stationary points, due to the sensitivity of the optimizer to scale (elaborated below). Thus we suggest that $\bm{\Sigma} = \mathbf{I}$ is likely a good choice for simple data, with lower values for more complex data.

\begin{figure}
\begin{subfigure}[h]{0.32\linewidth}
\includegraphics[width=\linewidth]{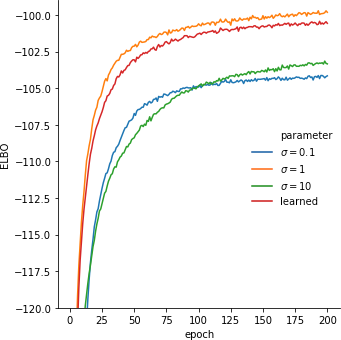}
\caption{Unmodified network.}\label{fig-sigmas}
\end{subfigure}
\hfil
\begin{subfigure}[h]{0.32\linewidth}
\includegraphics[width=\linewidth]{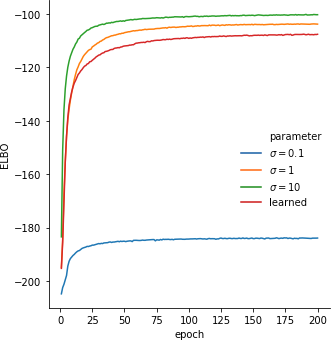}
\caption{Encoder $\times$ \num{10}.}\label{fig-sigmas-x10}
\end{subfigure}
\hfil
\begin{subfigure}[h]{0.32\linewidth}
\includegraphics[width=\linewidth]{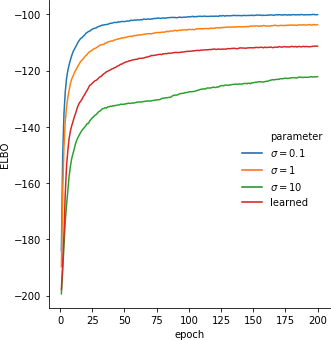}
\caption{Encoder $\times$ \num{0.1}.}\label{fig-sigmas-d10}
\end{subfigure}
\caption{
Progress of variational lower bound during training (mean of three runs) with hard coded scaling in the encoder and decoder to highlight the impact of initialization. Parameter $\sigma$ indicates BILBO using $\bm{\Sigma} = \sigma^2 \mathbf{I}$ (\cref{eq-bilbo}). ``learned'' indicates \cref{eq-vae} with learned $\bm{\Sigma}$.
Observe that the learned method is always most comparable to $\sigma = 1$ regardless of the scaling factor, and the best performing method is always the constant posterior variance that matches the scaling factor, suggesting that the difference is due to the initialization of network weights rather than the objective.
}\label{fig-sigma-ablation}
\end{figure}

We noted that the optimizer was sensitive to the scale of the data space and latent space, even in cases where the objective ought to have mathematically equivalent stationary points. \cref{fig-sigma-ablation} graphs the progress of the variational lower bound for several runs that include a hard coded scaling at the end of the encoder network (and inverse scaling at the beginning of the decoder network), and scaling the constant posterior variance by several different factors. All of these runs ought to have identical variational lower bound, by \cref{eq-vae-s}. We observe that the performance of the optimizer is best when the posterior variance is scaled commensurately with the hard coded scaling, suggesting a sensitivity to the initial conditions. Further, we observe that the learned posterior variance method always performs most similarly to (but a little worse than) the unscaled constant posterior variance, which validates that the higher performance of the proposed method is not merely due to lucky initialization.

\subsection{Experiments with Likelihood Variances}

We experimented on VAEs with Gaussian likelihood, using the BILBO objective (\cref{eq-bilbo}). We first trained a VAE with learned $\bm{\Tau}$, without additional constraints such as weight decay. These experiments did not converge: the ELBO fluctuated up and down, the latent vector scatter plots resembled a whirling flock of starlings, and generative samples loosely resemble the mean (see \cref{fig-unbounded-tau}, top row). We next trained VAEs with constant $\bm{\Tau} = \mathbf{I}$, and artificially scaled the data vectors to test the sensitivity to changes in the scale of the data. \cref{fig-unbounded-tau} illustrates that a modest change in the scale produces a dramatically different generative sample quality.
We then repeated similar experiments, but using BAGGINS to determine $\bm{\Tau}$ (\cref{eq-baggins}). \cref{fig-generative-baggins-consistency} illustrates that BAGGINS produces generative samples that are insensitive to the scale of the data. We selected the information factor $\tau$ in \cref{eq-baggins} by performing a parameter sweep and selecting one by inspection. \cref{fig-generative-baggins-tau} illustrates the effect of different values of $\tau$, with lower values appearing somewhat blurry and higher values breaking apart.
We note that generative samples using Bernoulli likelihood are more plausible than Gaussian likelihood for MNIST, perhaps better modeling the perceptual distance for pen and ink. We also note that this and other effects are most obvious for high latent dimensionality, perhaps due to the bottleneck of low latent dimensionality further regularizing the space. (See \cref{fig-baggins-scatter-lambda,fig-baggins-scatter-tau,fig-constant-manifold,fig-baggins-manifold} for additional results.)

\section{Limitations and Future Work}

Our analysis of the ELBO, and our proposed BILBO formulation, assume Gaussian prior and posterior. Likewise, our proposed BAGGINS formulation assumes Gaussian likelihoods. It may be possible to extend this work to other families of distributions and likelihoods (such as perceptual losses or discriminative losses). Developing this into a general method may enable simpler, more broadly applicable training procedures that are less sensitive to hyperparameter tuning.

In \cref{sec-discussion} we discuss how the blurriness typical to VAEs actually results from the choice of loss function rather than a fundamental limitation of VAEs. Developing loss functions that induce sharp interpolations would be a valuable contribution broadly applicable to the field.

We relate the exact log evidence for noisy generative models in \cref{eq-exact-noisy-general,eq-exact-noisy} to normalizing flow networks. It would be interesting to employ these equations to introduce noise modeling into normalizing flow techniques, which ordinarily assume a noise-free model.

Finally, the stochastic estimator for the log determinant using only traces in \cref{eq-trace-log} (and its particular appearance in the ELBO as \cref{eq-sigma-role}) is to our knowledge a novel estimator that may find further applications in the field of deep learning or data modeling more broadly.

\section{Conclusions}
\vspace{-1ex}
We have shown through argument and experiment that the posterior variances (in the encoder) in a variational auto-encoder may simply be constant without detriment to the optimal variational lower bound, in the case of Gaussian prior and posterior. This leads to a simplified reformulation for training VAEs in batches, which we call the Batch Information Lower Bound (BILBO).
We have also shown that the variance for Gaussian likelihood (in the decoder) is optimal in the limit approaching zero, which is unstable without regularization. We have proposed a bounded solution based on a presumed information content in the data population, and an associated estimator for training in batches, which we call Bounded Aggregate Information Sampling (BAGGINS).

While our experiments employ basic network architectures and data sets to validate our theoretical results, we hope that our exposition and supporting contributions provide useful insight into the action of variational auto-encoders, which remain popular in recent state of the art literature.

The next time you use a VAE in a project, you'll know that the learned variances can be omitted. But you may also benefit from knowing that the ELBO is exact log evidence at its stationary points, or that your loss function is what's making your results blurry; not the VAE.

\section*{Broader Impact}

Our work is primarily theoretical, but may simplify the development of new applications for auto-encoders, or improve the efficiency of existing applications. It is notable that one application of auto-encoders with some controversy is the so-called "deep fakes". However, many beneficial applications exist including medical imaging, data compression, anomaly detection, recommendation systems, and generating content for entertainment.

\begin{ack}
\vspace{-1ex}
We wish to thank Jon Barron, Michael Broxton, Paul Debevec, John Flynn, and Matt Hoffman.
\end{ack}

\vspace{-1ex}

\printbibliography

\appendix

\section{Derivative of Expected Log Likelihood w.r.t.\ Posterior Mean}\label{appendix-dmu}

We expand the derivative of the expected log likelihood with respect to $\bm{\mu}$ as:
\begin{multline}
    \pder{}{\bm{\mu}}
    \mathbb{E}_{p(\mathbf{x})}\! \left[ \mathbb{E}_{\mathcal{N}(\mathbf{z}; \bm{\mu}, \bm{\Sigma})} [
    \log p_{\bm{\theta}}(\mathbf{x}|\mathbf{z}) ] \right]
    =
    -\pder{}{\bm{\mu}}
    \tfrac{1}{2N}\! \sum_i \int_{\mathbf{z}}\!
    \mathcal{N}(\mathbf{z}; \bm{\mu}_i, \bm{\Sigma}_i)
    |\bm{\nu}(\mathbf{z}) - \mathbf{x}_i|^2 d\mathbf{z}
    \\
    =
    -\tfrac{1}{2N}\! \sum_i \int_{\mathbf{z}}\! \big(
    |\bm{\nu}(\mathbf{z}) - \mathbf{x}_i|^2
    \pder{}{\bm{\mu}}
    \mathcal{N}(\mathbf{z}; \bm{\mu}_i, \bm{\Sigma}_i)
    +
    \mathcal{N}(\mathbf{z}; \bm{\mu}_i, \bm{\Sigma}_i)
    \pder{}{\bm{\mu}}
    |\bm{\nu}(\mathbf{z}) - \mathbf{x}_i|^2 \big) d\mathbf{z}
    \\
    =
    -\tfrac{1}{2N}\! \int_{\mathbf{z}}\! \big(
    |\bm{\nu}(\mathbf{z}) - \mathbf{x}|^2
    \pder{}{\bm{\mu}} \mathcal{N}(\mathbf{z}; \bm{\mu}, \bm{\Sigma})
    + 2 \sum_i 
    \mathcal{N}(\mathbf{z}; \bm{\mu}_i, \bm{\Sigma}_i)
    (\bm{\nu}(\mathbf{z}) - \mathbf{x}_i)^\tran\!
    \pder{}{\bm{\mu}}
    \bm{\nu}(\mathbf{z}) \big) d\mathbf{z}
    \\
    =
    -\tfrac{1}{2N}\! \int_{\mathbf{z}}
    |\bm{\nu}(\mathbf{z}) - \mathbf{x}|^2
    \mathcal{N}(\mathbf{z}; \bm{\mu}, \bm{\Sigma})
    (\mathbf{z} - \bm{\mu})^\tran \bm{\Sigma}^\inv d\mathbf{z}.
    \label{eq-ap-dmu-1}
\end{multline}
Note that the summation including the $(\bm{\nu}(\mathbf{z}) - \mathbf{x}_i)$ term vanishes due to the solution of $\bm{\nu}(\mathbf{z})$.
We may change the expectation in \cref{eq-ap-dmu-1} to standard normal domain as follows:
\begin{equation}
    \pder{}{\bm{\mu}}
    \mathbb{E}_{p(\mathbf{x})}\! \left[ \mathbb{E}_{\mathcal{N}(\mathbf{z}; \bm{\mu}, \bm{\Sigma})} [
    \log p_{\bm{\theta}}(\mathbf{x}|\mathbf{z}) ] \right]
    =
    -\tfrac{1}{2N} \mathbb{E}_{\mathcal{N}(\bm{\epsilon}; \mathbf{0}, \mathbf{I})}\! \big[
    |\bm{\nu}(\bm{\mu} + \sqrt{\bm{\Sigma}} \bm{\epsilon}) - \mathbf{x}|^2
    \bm{\epsilon}^\tran \bm{\Sigma}^\invsqrt \big].
\end{equation}
Let $\mathbf{J}$ be defined such that $\bm{\nu}(\bm{\mu} + \sqrt{\bm{\Sigma}} \bm{\epsilon}) \simeq \bm{\nu}(\bm{\mu}) + \mathbf{J} \sqrt{\bm{\Sigma}} \bm{\epsilon}$ over $\bm{\epsilon} \sim \mathcal{N}(\mathbf{0}, \mathbf{I})$; in other words a local linear fit windowed by the posterior. By the smoothness of $\bm{\nu}$ we expect the error of this approximation to be small. Referring to \cite{IMM2012-03274} we recall\footnote{\cite{IMM2012-03274} Eq.~(394) contains an error; the expression is transposed.} the identities
$\mathbb{E}_{\mathcal{N}(\mathbf{v}; \mathbf{0}, \mathbf{I})} [\mathbf{v}^\tran] = \mathbf{0}$,
$\mathbb{E}_{\mathcal{N}(\mathbf{v}; \mathbf{0}, \mathbf{I})} [\mathbf{A} \mathbf{v} \mathbf{v}^\tran] = \mathbf{A}$,
and
$\mathbb{E}_{\mathcal{N}(\mathbf{v}; \mathbf{0}, \mathbf{I})} [\mathbf{v}^\tran\! \mathbf{A} \mathbf{v} \mathbf{v}^\tran] = \mathbf{0}$ and perform the following manipulations:
\begin{multline}
    -2N \pder{}{\bm{\mu}}
    \mathbb{E}_{p(\mathbf{x})}\! \left[ \mathbb{E}_{\mathcal{N}(\mathbf{z}; \bm{\mu}, \bm{\Sigma})} [
    \log p_{\bm{\theta}}(\mathbf{x}|\mathbf{z}) ] \right]
    \! \sqrt{\bm{\Sigma}}
    =
    \mathbb{E}_{\mathcal{N}(\bm{\epsilon}; \mathbf{0}, \mathbf{I})}\! \big[
    |\bm{\nu}(\bm{\mu} + \sqrt{\bm{\Sigma}} \bm{\epsilon}) - \mathbf{x}|^2
    \bm{\epsilon}^\tran \big]
    \\
    \simeq
    \mathbb{E}_{\mathcal{N}(\bm{\epsilon}; \mathbf{0}, \mathbf{I})}\! \big[
    |\bm{\nu}(\bm{\mu}) + \mathbf{J} \sqrt{\bm{\Sigma}} \bm{\epsilon} - \mathbf{x}|^2
    \bm{\epsilon}^\tran \big]
    \\
    =
    \mathbb{E}_{\mathcal{N}(\bm{\epsilon}; \mathbf{0}, \mathbf{I})}\! \big[
    |\bm{\nu}(\bm{\mu}) - \mathbf{x}|^2 \bm{\epsilon}^\tran
    + 2 (\bm{\nu}(\bm{\mu}) - \mathbf{x})^\tran \mathbf{J} \sqrt{\bm{\Sigma}} \bm{\epsilon} \bm{\epsilon}^\tran
    + \bm{\epsilon}^\tran \sqrt{\bm{\Sigma}}^\tran \mathbf{J}^\tran\!
    \mathbf{J} \sqrt{\bm{\Sigma}} \bm{\epsilon}
    \bm{\epsilon}^\tran \big]
    \\
    =
    2 (\bm{\nu}(\bm{\mu}) - \mathbf{x})^\tran \mathbf{J} \sqrt{\bm{\Sigma}}.
\end{multline}
Therefore we may state:
\begin{equation}
    \pder{}{\bm{\mu}}
    \mathbb{E}_{p(\mathbf{x})}\! \left[ \mathbb{E}_{\mathcal{N}(\mathbf{z}; \bm{\mu}, \bm{\Sigma})} [
    \log p_{\bm{\theta}}(\mathbf{x}|\mathbf{z}) ] \right]
    \simeq
    -\tfrac{1}{N} (\bm{\nu}(\bm{\mu}) - \mathbf{x})^\tran \mathbf{J}.
    \label{eq-ap-dmu-3}
\end{equation}

\pagebreak

\section{Derivative of Expected Log Likelihood w.r.t.\ Posterior Variance}\label{appendix-dsigma}

We expand the derivative of the expected log likelihood with respect to $\bm{\Sigma}$ as:
\begin{multline}
    \pder{}{\bm{\Sigma}}
    \mathbb{E}_{p(\mathbf{x})}\! \left[ \mathbb{E}_{\mathcal{N}(\mathbf{z}; \bm{\mu}, \bm{\Sigma})} [
    \log p_{\bm{\theta}}(\mathbf{x}|\mathbf{z}) ] \right]
    =
    -\pder{}{\bm{\Sigma}}
    \tfrac{1}{2N}\! \sum_i \int_{\mathbf{z}}\!
    \mathcal{N}(\mathbf{z}; \bm{\mu}_i, \bm{\Sigma}_i)
    |\bm{\nu}(\mathbf{z}) - \mathbf{x}_i|^2 d\mathbf{z}
    \\
    =
    -\tfrac{1}{2N}\! \sum_i \int_{\mathbf{z}}\! \big(
    |\bm{\nu}(\mathbf{z}) - \mathbf{x}_i|^2
    \pder{}{\bm{\Sigma}}
    \mathcal{N}(\mathbf{z}; \bm{\mu}_i, \bm{\Sigma}_i)
    +
    \mathcal{N}(\mathbf{z}; \bm{\mu}_i, \bm{\Sigma}_i)
    \pder{}{\bm{\Sigma}}
    |\bm{\nu}(\mathbf{z}) - \mathbf{x}_i|^2 \big) d\mathbf{z}
    \\
    =
    -\tfrac{1}{2N}\! \int_{\mathbf{z}}\! \big(
    |\bm{\nu}(\mathbf{z}) - \mathbf{x}|^2
    \pder{}{\bm{\Sigma}} \mathcal{N}(\mathbf{z}; \bm{\mu}, \bm{\Sigma})
    - 2 \sum_i 
    \mathcal{N}(\mathbf{z}; \bm{\mu}_i, \bm{\Sigma}_i)
    (\mathbf{x}_i - \bm{\nu}(\mathbf{z}))^\tran\!
    \otimes
    \pder{}{\bm{\Sigma}}
    \bm{\nu}(\mathbf{z}) \big) d\mathbf{z}
    \\
    =
    -\tfrac{1}{4N}\! \int_{\mathbf{z}}
    |\bm{\nu}(\mathbf{z}) - \mathbf{x}|^2
    \mathcal{N}(\mathbf{z}; \bm{\mu}, \bm{\Sigma})
    (\bm{\Sigma}^\inv (\bm{\mu} - \mathbf{z}) (\bm{\mu} - \mathbf{z})^\tran \bm{\Sigma}^\inv - \bm{\Sigma}^\inv) d\mathbf{z}.
    \label{eq-ap-dsigma-2}
\end{multline}
Note that the summation including the $(\bm{\nu}(\mathbf{z}) - \mathbf{x}_i)$ term vanishes due to the solution of $\bm{\nu}(\mathbf{z})$.
We may change the expectation in \cref{eq-ap-dsigma-2} to standard normal domain as follows:
\begin{multline}
    \pder{}{\bm{\Sigma}}
    \mathbb{E}_{p(\mathbf{x})}\! \left[ \mathbb{E}_{\mathcal{N}(\mathbf{z}; \bm{\mu}, \bm{\Sigma})} [
    \log p_{\bm{\theta}}(\mathbf{x}|\mathbf{z}) ] \right] \\
    =
    -\tfrac{1}{4N} \mathbb{E}_{\mathcal{N}(\bm{\epsilon}; \mathbf{0}, \mathbf{I})}\! \big[
    |\bm{\nu}(\bm{\mu} + \sqrt{\bm{\Sigma}} \bm{\epsilon}) - \mathbf{x}|^2
    (\bm{\Sigma}^\invsqrt \bm{\epsilon} \bm{\epsilon}^\tran \bm{\Sigma}^\itransqrt - \bm{\Sigma}^\inv) \big].
\end{multline}
Let $\mathbf{J}$ be defined such that $\bm{\nu}(\bm{\mu} + \sqrt{\bm{\Sigma}} \bm{\epsilon}) \simeq \bm{\nu}(\bm{\mu}) + \mathbf{J} \sqrt{\bm{\Sigma}} \bm{\epsilon}$ over $\bm{\epsilon} \sim \mathcal{N}(\mathbf{0}, \mathbf{I})$; in other words a local linear fit windowed by the posterior. By the smoothness of $\bm{\nu}$ we expect the error of this approximation to be small. Referring to \cite{IMM2012-03274} we recall\footnote{\cite{IMM2012-03274} Eq.~(348) contains an error; a factor of one half is missing (though present in the related Eq.~(396)).} the identities
$\mathbb{E}_{\mathcal{N}(\mathbf{v}; \mathbf{0}, \mathbf{I})} [\mathbf{A} \mathbf{v}] = \mathbf{0}$,
$\mathbb{E}_{\mathcal{N}(\mathbf{v}; \mathbf{0}, \mathbf{I})} [\mathbf{v} \mathbf{v}^\tran] = \mathbf{I}$,
$\mathbb{E}_{\mathcal{N}(\mathbf{v}; \mathbf{0}, \mathbf{I})} [\mathbf{v}^\tran\! \mathbf{A} \mathbf{v}] = \tr({\mathbf{A}})$,
$\mathbb{E}_{\mathcal{N}(\mathbf{v}; \mathbf{0}, \mathbf{I})} [\mathbf{v}^\tran\! \mathbf{A} \mathbf{v} \mathbf{v}^\tran] = \mathbf{0}$,
and
$\mathbb{E}_{\mathcal{N}(\mathbf{v}; \mathbf{0}, \mathbf{I})} [\mathbf{v} \mathbf{v}^\tran\! \mathbf{A} \mathbf{v} \mathbf{v}^\tran] = \mathbf{A} + \mathbf{A}^\tran + \tr({\mathbf{A}}) \mathbf{I}$ and perform the following manipulations:
\begin{multline}
    -4N \sqrt{\bm{\Sigma}}^\tran\!
    \pder{}{\bm{\Sigma}}
    \mathbb{E}_{p(\mathbf{x})}\! \left[ \mathbb{E}_{\mathcal{N}(\mathbf{z}; \bm{\mu}, \bm{\Sigma})} [
    \log p_{\bm{\theta}}(\mathbf{x}|\mathbf{z}) ] \right]
    \!\sqrt{\bm{\Sigma}}
    =
    \mathbb{E}_{\mathcal{N}(\bm{\epsilon}; \mathbf{0}, \mathbf{I})}\! \big[
    |\bm{\nu}(\bm{\mu} + \sqrt{\bm{\Sigma}} \bm{\epsilon}) - \mathbf{x}|^2
    (\bm{\epsilon} \bm{\epsilon}^\tran - \mathbf{I}) \big]
    \\
    \simeq
    \mathbb{E}_{\mathcal{N}(\bm{\epsilon}; \mathbf{0}, \mathbf{I})}\! \big[
    |\bm{\nu}(\bm{\mu}) + \mathbf{J} \sqrt{\bm{\Sigma}} \bm{\epsilon} - \mathbf{x}|^2
    (\bm{\epsilon} \bm{\epsilon}^\tran - \mathbf{I}) \big]
    \\
    =
    \mathbb{E}_{\mathcal{N}(\bm{\epsilon}; \mathbf{0}, \mathbf{I})}\! \big[
    \big(
    |\bm{\nu}(\bm{\mu}) - \mathbf{x}|^2
    + 2 (\bm{\nu}(\bm{\mu}) - \mathbf{x})^\tran \mathbf{J} \sqrt{\bm{\Sigma}} \bm{\epsilon}
    + |\mathbf{J} \sqrt{\bm{\Sigma}} \bm{\epsilon}|^2
    \big)
    (\bm{\epsilon} \bm{\epsilon}^\tran - \mathbf{I}) \big]
    \\
    =
    \mathbb{E}_{\mathcal{N}(\bm{\epsilon}; \mathbf{0}, \mathbf{I})}\! \big[
    |\bm{\nu}(\bm{\mu}) - \mathbf{x}|^2
    (\bm{\epsilon} \bm{\epsilon}^\tran - \mathbf{I})
    + 2 \bm{\epsilon}^\tran \sqrt{\bm{\Sigma}}^\tran\! \mathbf{J}^\tran
    (\bm{\nu}(\bm{\mu}) - \mathbf{x})
    \bm{\epsilon} \bm{\epsilon}^\tran \\
    - (2 (\bm{\nu}(\bm{\mu}) - \mathbf{x})^\tran
    \mathbf{J} \sqrt{\bm{\Sigma}} \bm{\epsilon}) \mathbf{I}
    + \bm{\epsilon} \bm{\epsilon}^\tran\! \sqrt{\bm{\Sigma}}^\tran\! \mathbf{J}^\tran\! \mathbf{J} \sqrt{\bm{\Sigma}} \bm{\epsilon} \bm{\epsilon}^\tran - |\mathbf{J} \sqrt{\bm{\Sigma}} \bm{\epsilon}|^2\mathbf{I}
    \big]
    \\
    =
    |\bm{\nu}(\bm{\mu}) - \mathbf{x}|^2
    (\mathbf{I} - \mathbf{I})
    + 2 \sqrt{\bm{\Sigma}}^\tran\! \mathbf{J}^\tran\! \mathbf{J} \sqrt{\bm{\Sigma}} + \tr(\sqrt{\bm{\Sigma}}^\tran\! \mathbf{J}^\tran\! \mathbf{J} \sqrt{\bm{\Sigma}})\mathbf{I} - \tr(\sqrt{\bm{\Sigma}}^\tran\! \mathbf{J}^\tran\! \mathbf{J} \sqrt{\bm{\Sigma}})\mathbf{I}
    \\
    =
    2 \sqrt{\bm{\Sigma}}^\tran\! \mathbf{J}^\tran\! \mathbf{J} \sqrt{\bm{\Sigma}}.
\end{multline}
Therefore we may state:
\begin{equation}
    \pder{}{\bm{\Sigma}}
    \mathbb{E}_{p(\mathbf{x})}\! \left[ \mathbb{E}_{\mathcal{N}(\mathbf{z}; \bm{\mu}, \bm{\Sigma})} [
    \log p_{\bm{\theta}}(\mathbf{x}|\mathbf{z}) ] \right]
    \simeq
    -\tfrac{1}{2N} \mathbf{J}^\tran\! \mathbf{J}.
    \label{eq-ap-dsigma-3}
\end{equation}

\pagebreak

\clearpage


\section{Additional Figures}\label{appendix-figures}

\begin{figure}[ht]
\begin{subfigure}[h]{0.32\linewidth}
\includegraphics[width=\linewidth]{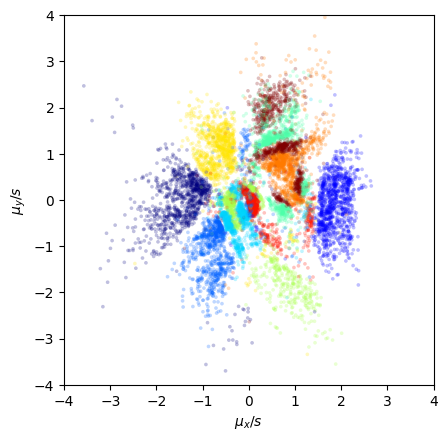}
\caption{$\bm{\Sigma} = 0.1$.}
\end{subfigure}
\hfil
\begin{subfigure}[h]{0.32\linewidth}
\includegraphics[width=\linewidth]{figures/scatter_mu_bilbo}
\caption{$\bm{\Sigma} = 1$.}
\end{subfigure}
\hfil
\begin{subfigure}[h]{0.32\linewidth}
\includegraphics[width=\linewidth]{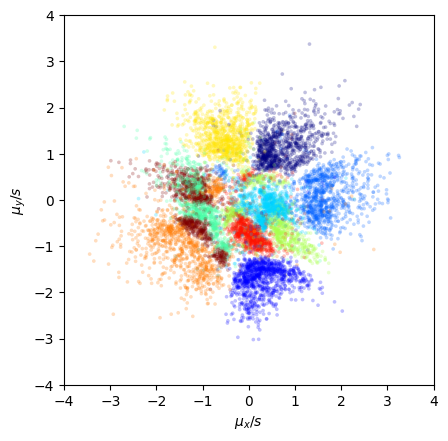}
\caption{$\bm{\Sigma} = 10$.}
\end{subfigure}
\caption{
Posterior means for a two-dimensional MNIST latent space colored by label, with Bernoulli likelihood, trained using BILBO with different values of $\bm{\Sigma}$. The distributions are qualitatively similar.
}\label{fig-bilbo-scatter-sigma}
\vspace{-0.25cm}
\end{figure}

\begin{figure}[ht]
\centering
\begin{subfigure}[h]{0.32\linewidth}
\includegraphics[width=\linewidth]{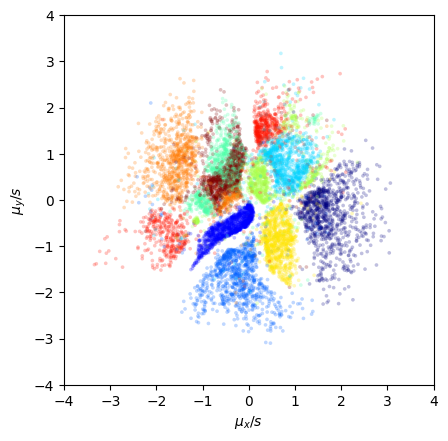}
\caption{$\lambda = 1$.}
\end{subfigure}
\hfil
\begin{subfigure}[h]{0.32\linewidth}
\includegraphics[width=\linewidth]{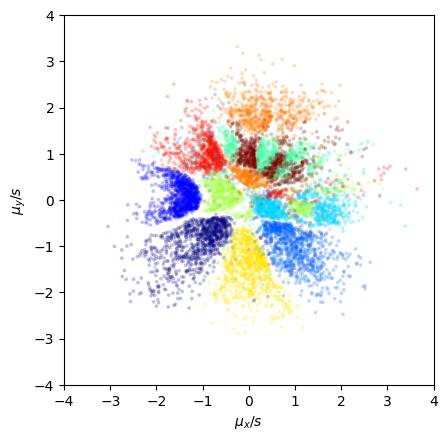}
\caption{$\lambda = 2$.}
\end{subfigure}
\hfil
\begin{subfigure}[h]{0.32\linewidth}
\includegraphics[width=\linewidth]{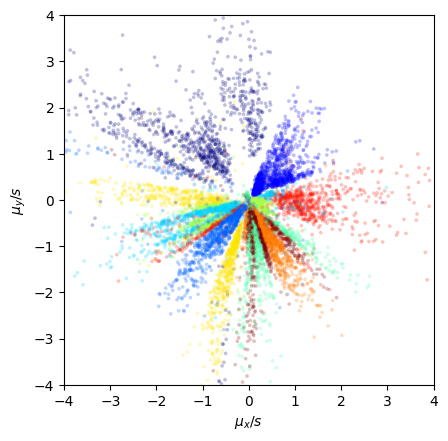}
\caption{$\lambda = 10$.}
\end{subfigure}
\caption{
Posterior means for a two-dimensional MNIST latent space colored by label, with Gaussian likelihood with constant variance $\bm{\Tau} = \mathbf{I}$. $\lambda$ indicates a scaling factor applied to the data vectors. Qualitatively, the distributions pinch towards the origin as $\lambda$ increases, becoming more radial.
}\label{fig-baggins-scatter-lambda}
\vspace{-0.25cm}
\end{figure}

\begin{figure}[ht]
\centering
\begin{subfigure}[h]{0.32\linewidth}
\includegraphics[width=\linewidth]{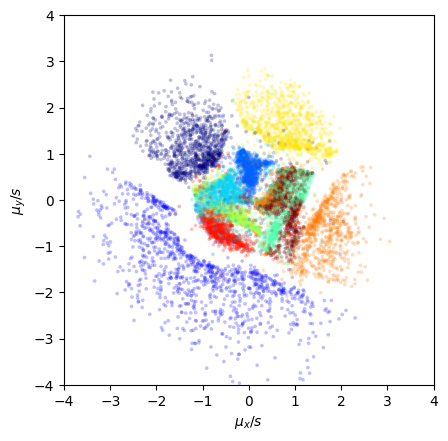}
\caption{$\tau = \nicefrac{1}{10}$.}
\end{subfigure}
\hfil
\begin{subfigure}[h]{0.32\linewidth}
\includegraphics[width=\linewidth]{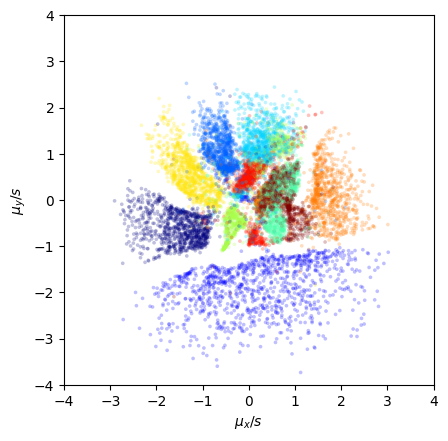}
\caption{$\tau = \nicefrac{1}{5}$.}
\end{subfigure}
\hfil
\begin{subfigure}[h]{0.32\linewidth}
\includegraphics[width=\linewidth]{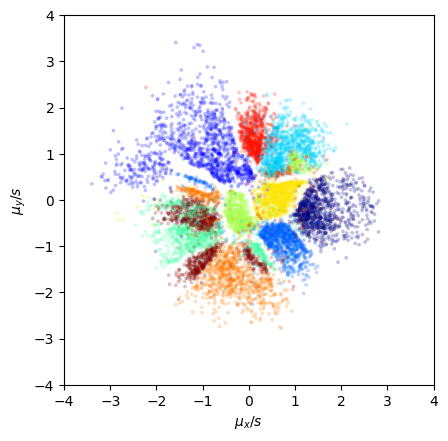}
\caption{$\tau = 1$.}
\end{subfigure}
\caption{
Posterior means for a two-dimensional MNIST latent space colored by label, with Gaussian likelihood using BAGGINS to determine likelihood variances with information factor $\tau$. Qualitatively, the gaps between label clusters tighten up as $\tau$ increases.
}\label{fig-baggins-scatter-tau}
\vspace{-0.25cm}
\end{figure}

\begin{figure}[ht]
\centering
\begin{subfigure}[h]{0.64\linewidth}
\includegraphics[width=\linewidth]{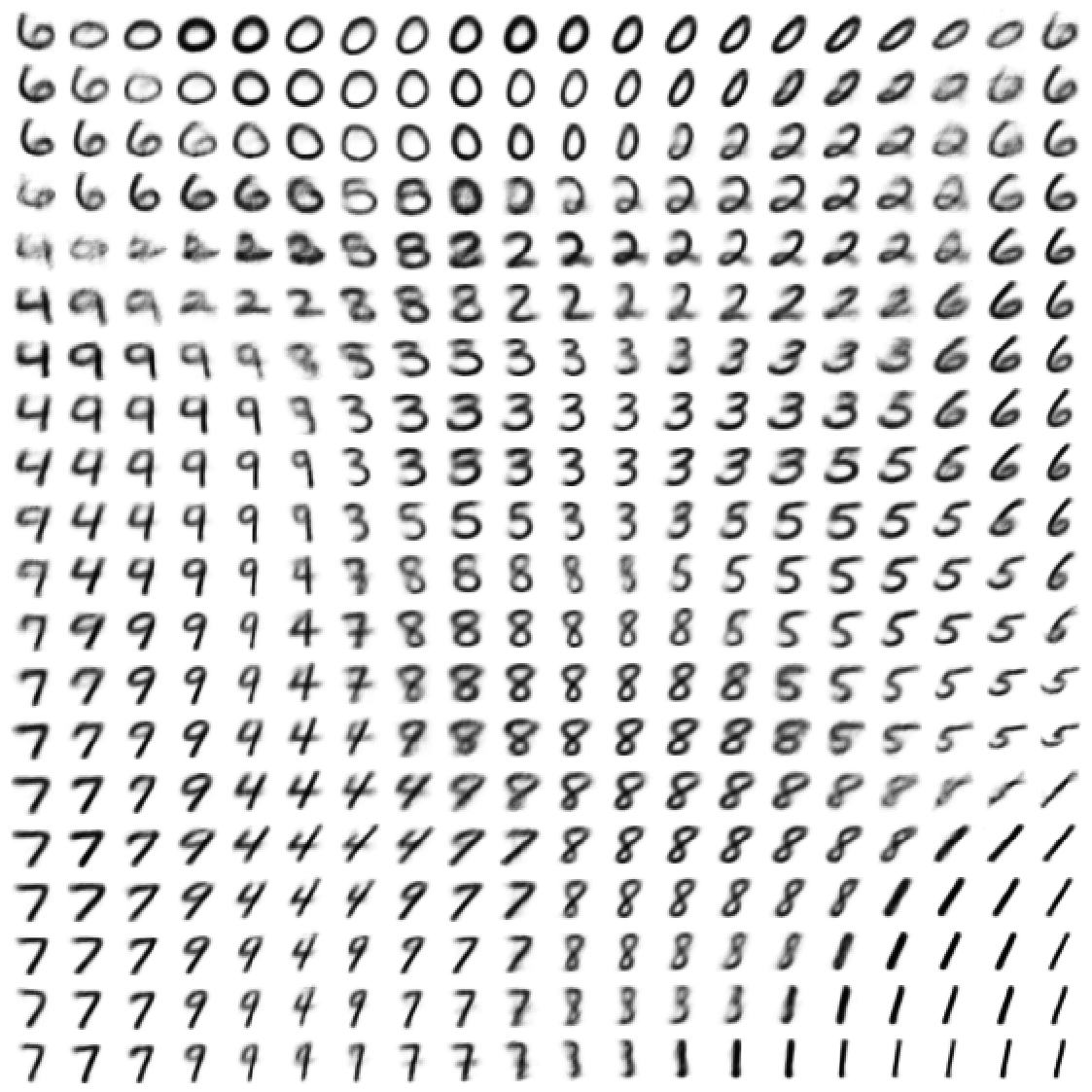}
\end{subfigure}
\hfil
\begin{subfigure}[h]{0.32\linewidth}
\includegraphics[width=\linewidth]{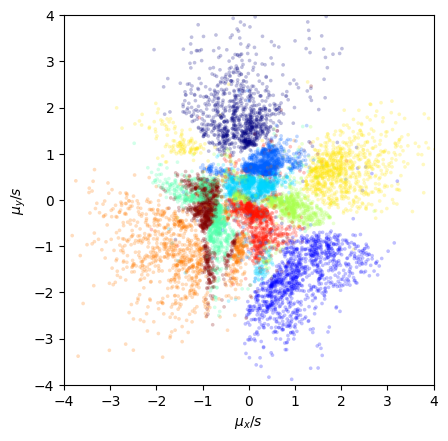}
\includegraphics[width=\linewidth]{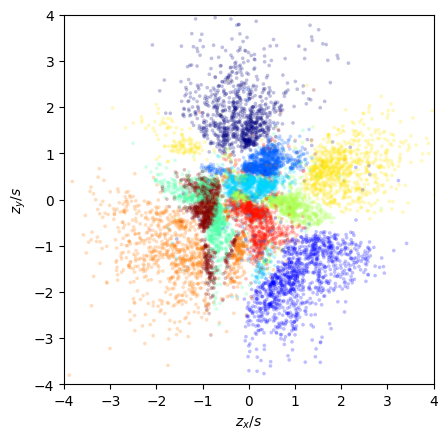}
\end{subfigure}
\caption{
Latent manifold and scatter plots for a two-dimensional MNIST latent space with Bernoulli likelihood and learned posterior variances. $\mathbf{z}$ scatter plot (lower right) is a single sample from the posterior, while $\bm{\mu}$ (upper right) is its mean.
}\label{fig-learned-manifold}
\vspace{-0.25cm}
\end{figure}

\begin{figure}[ht]
\centering
\begin{subfigure}[h]{0.64\linewidth}
\includegraphics[width=\linewidth]{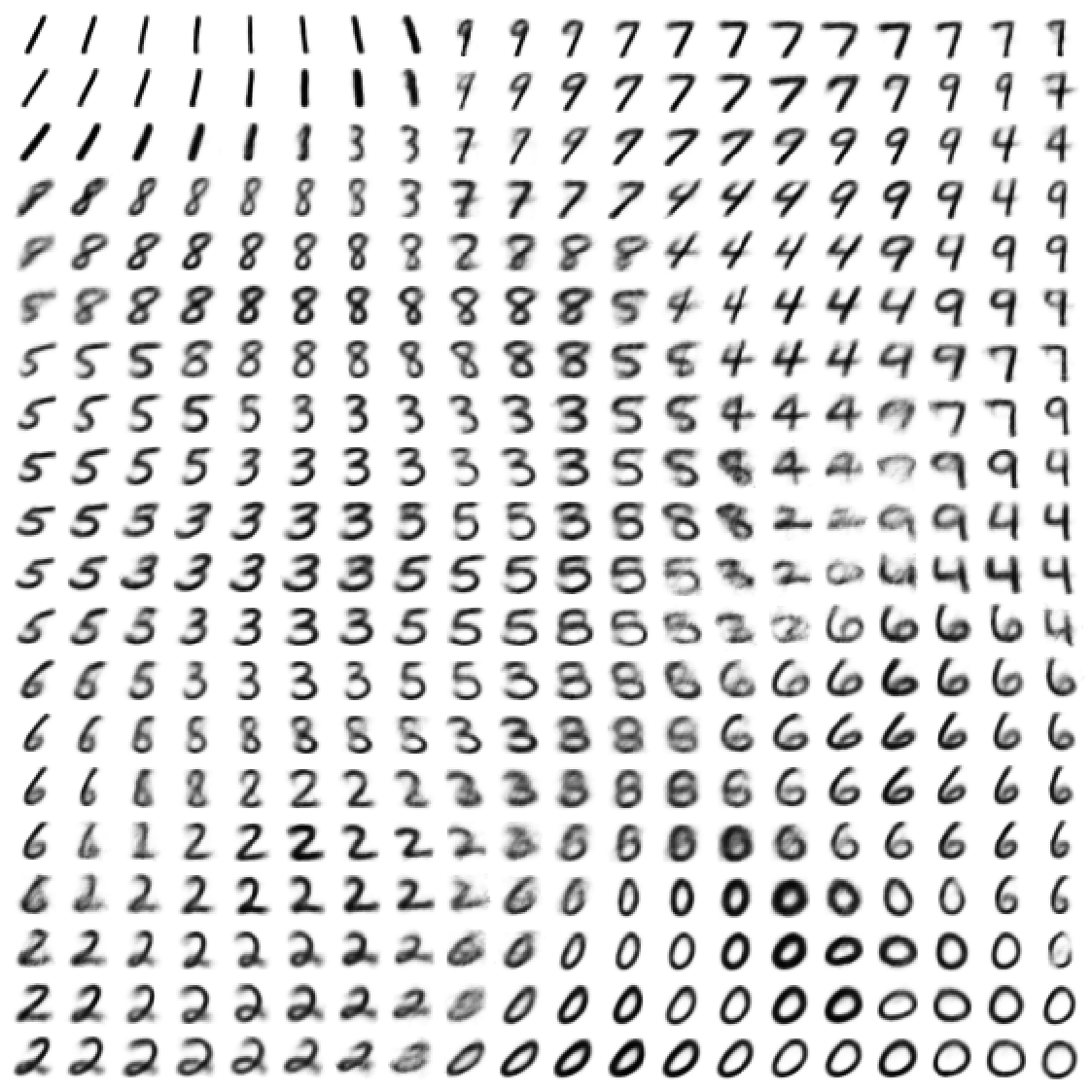}
\end{subfigure}
\hfil
\begin{subfigure}[h]{0.32\linewidth}
\includegraphics[width=\linewidth]{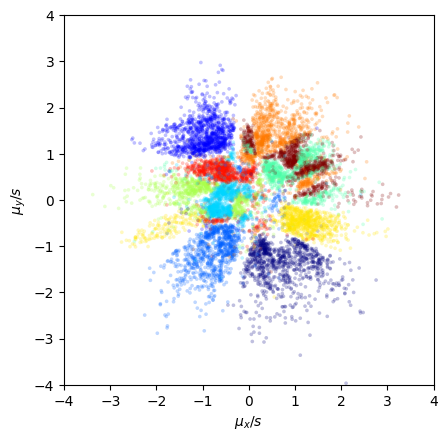}
\includegraphics[width=\linewidth]{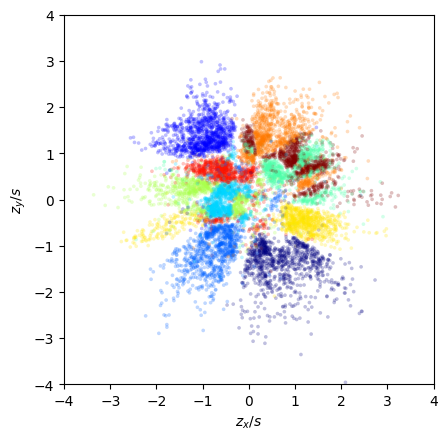}
\end{subfigure}
\caption{
Latent manifold and scatter plots for a two-dimensional MNIST latent space with Bernoulli likelihood, trained using BILBO with constant posterior variance $\bm{\Sigma} = \mathbf{I}$. $\mathbf{z}$ scatter plot (lower right) is a single sample from the posterior, while $\bm{\mu}$ (upper right) is its mean.
}\label{fig-bilbo-manifold}
\vspace{-0.25cm}
\end{figure}

\begin{figure}[ht]
\centering
\begin{subfigure}[h]{0.64\linewidth}
\includegraphics[width=\linewidth]{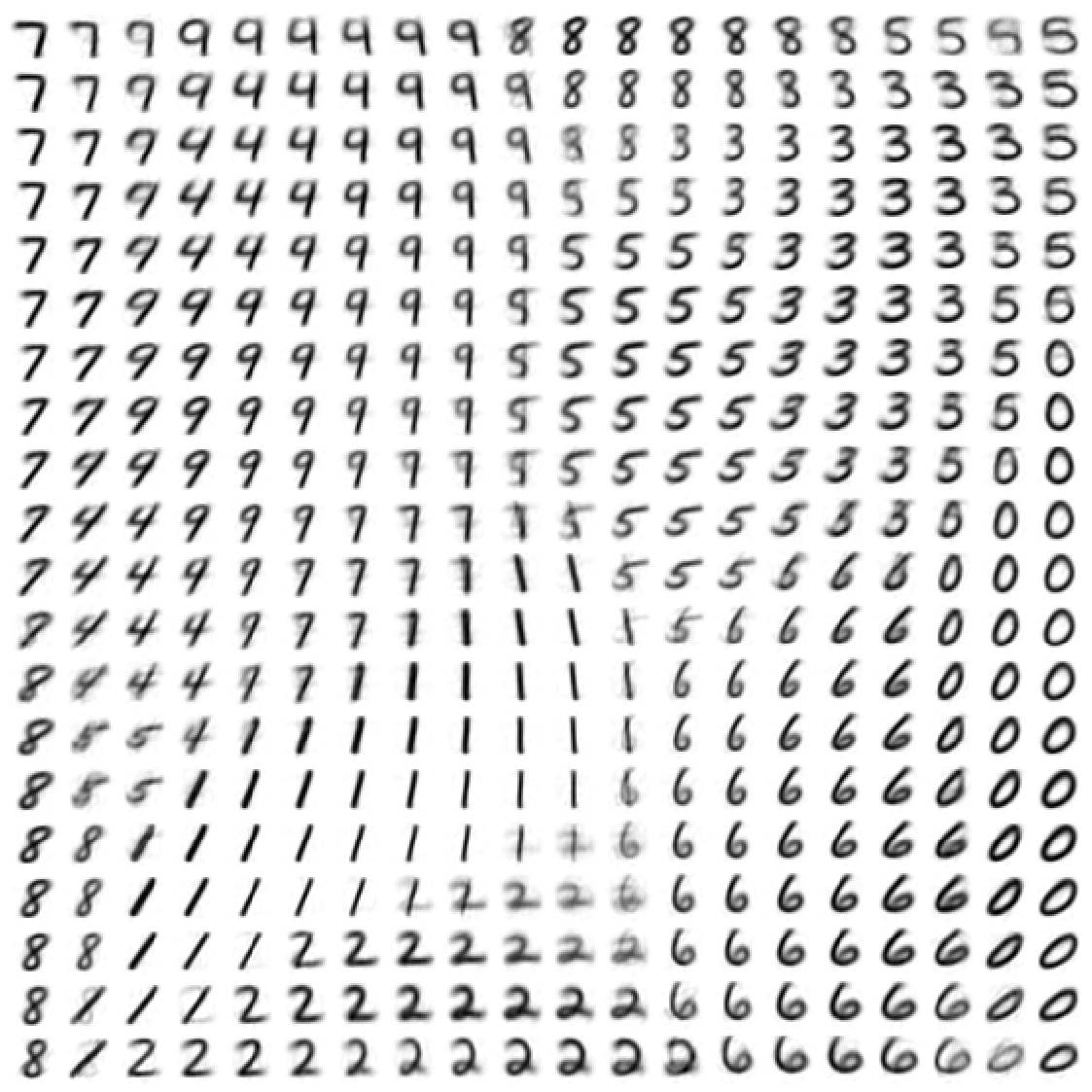}
\end{subfigure}
\hfil
\begin{subfigure}[h]{0.32\linewidth}
\includegraphics[width=\linewidth]{figures/scatter_mu_1_likelihood}
\includegraphics[width=\linewidth]{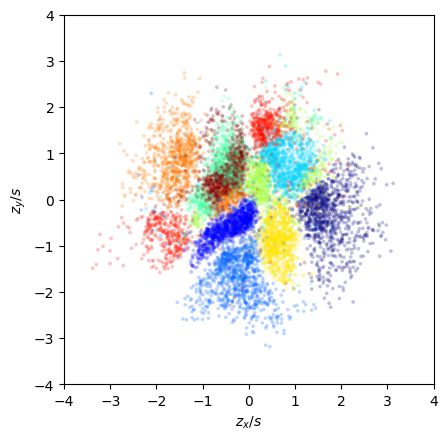}
\end{subfigure}
\caption{
Latent manifold and scatter plots for a two-dimensional MNIST latent space with Gaussian likelihood using constant variance $\bm{\Tau} = \mathbf{I}$. $\mathbf{z}$ scatter plot (lower right) is a single sample from the posterior, while $\bm{\mu}$ (upper right) is its mean.
}\label{fig-constant-manifold}
\vspace{-0.25cm}
\end{figure}

\begin{figure}[ht]
\centering
\begin{subfigure}[h]{0.64\linewidth}
\includegraphics[width=\linewidth]{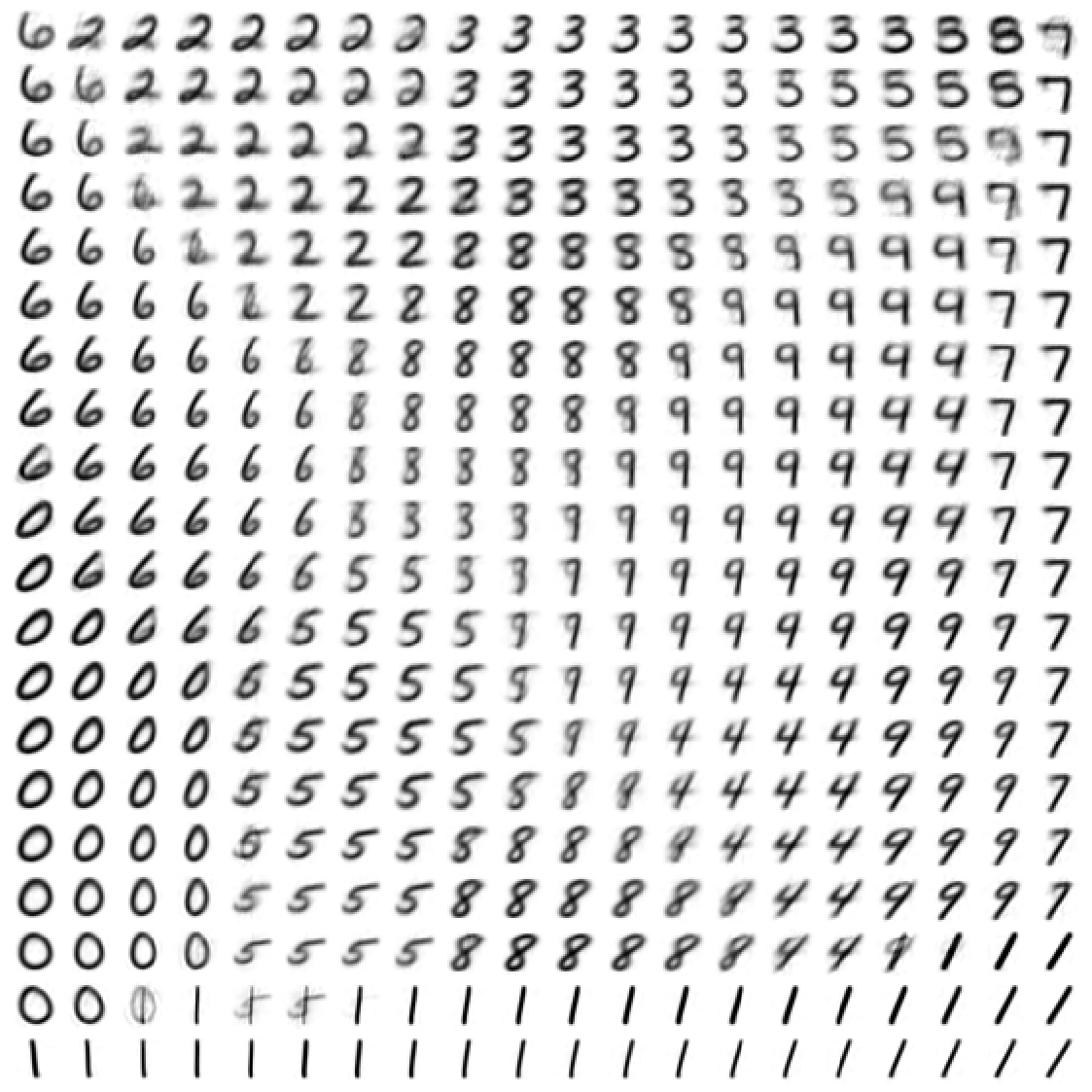}
\end{subfigure}
\hfil
\begin{subfigure}[h]{0.32\linewidth}
\includegraphics[width=\linewidth]{figures/scatter_mu_baggins_t02}
\includegraphics[width=\linewidth]{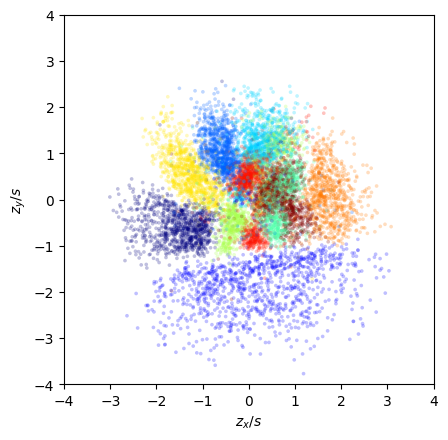}
\end{subfigure}
\caption{
Latent manifold and scatter plots for a two-dimensional MNIST latent space with Gaussian likelihood using BAGGINS with information factor $\tau = 1$. $\mathbf{z}$ scatter plot (lower right) is a single sample from the posterior, while $\bm{\mu}$ (upper right) is its mean.
}\label{fig-baggins-manifold}
\vspace{-0.25cm}
\end{figure}

\begin{figure}[ht]
\centering
\begin{subfigure}[h]{0.03\linewidth}\caption{}\end{subfigure}
\begin{subfigure}[h]{0.94\linewidth}
\includegraphics[width=\linewidth]{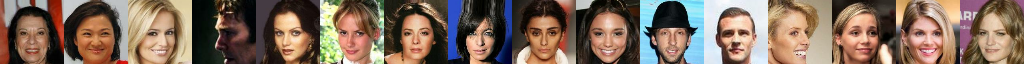}
\end{subfigure}
\begin{subfigure}[h]{0.03\linewidth}\caption{}\end{subfigure}
\begin{subfigure}[h]{0.94\linewidth}
\includegraphics[width=\linewidth]{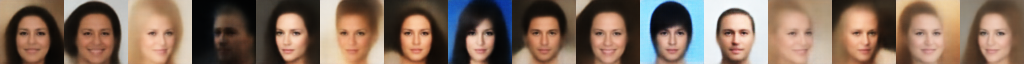}
\end{subfigure}
\begin{subfigure}[h]{0.03\linewidth}\caption{}\end{subfigure}
\begin{subfigure}[h]{0.94\linewidth}
\includegraphics[width=\linewidth]{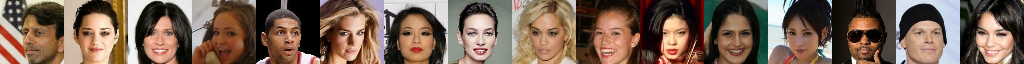}
\end{subfigure}
\begin{subfigure}[h]{0.03\linewidth}\caption{}\end{subfigure}
\begin{subfigure}[h]{0.94\linewidth}
\includegraphics[width=\linewidth]{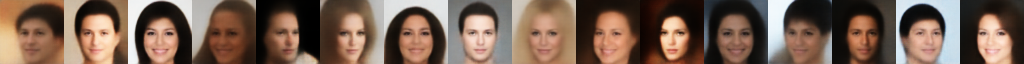}
\end{subfigure}
\caption{
Reconstructions for input data (a, c) using a VAE trained with (b) standard practice using ELBO and (d) BILBO. The two methods produce qualitatively similar reconstructions.
}\label{fig-celeb-recon}
\vspace{-0.25cm}
\end{figure}

\begin{figure}[ht]
\centering
\begin{subfigure}[h]{0.48\linewidth}
\includegraphics[width=\linewidth]{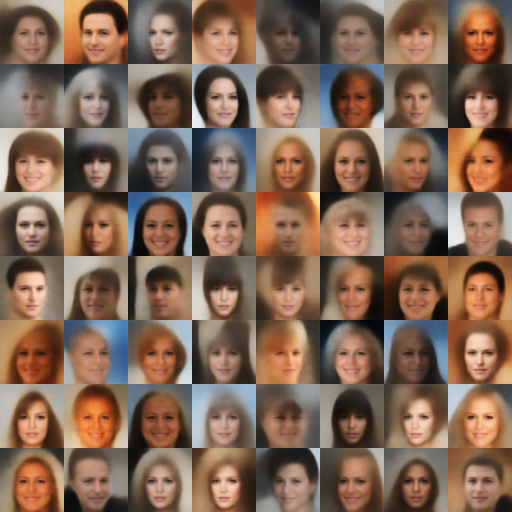}
\caption{Samples from standard normal prior; ELBO loss.}
\end{subfigure}
\hfil
\begin{subfigure}[h]{0.48\linewidth}
\includegraphics[width=\linewidth]{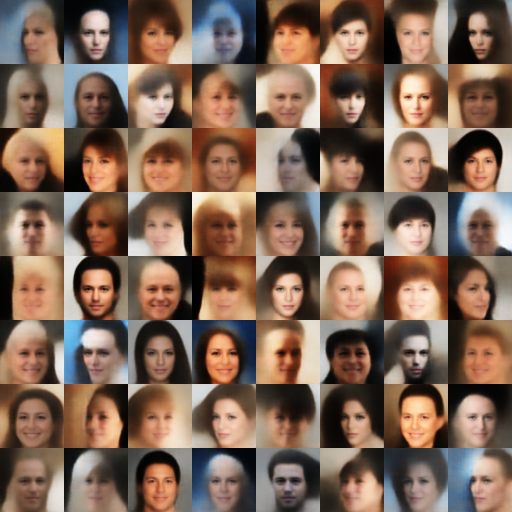}
\caption{Samples from optimal normal prior; ELBO loss.}
\end{subfigure}
\begin{subfigure}[h]{0.48\linewidth}
\includegraphics[width=\linewidth]{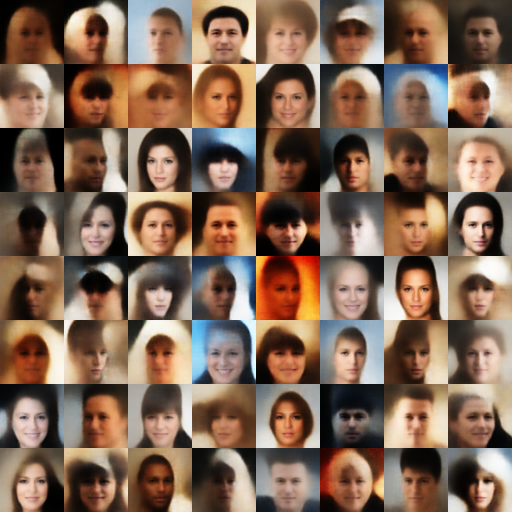}
\caption{Samples from prior; BILBO loss with $\bm{\Sigma} = \mathbf{I}$.}
\end{subfigure}
\hfil
\begin{subfigure}[h]{0.48\linewidth}
\includegraphics[width=\linewidth]{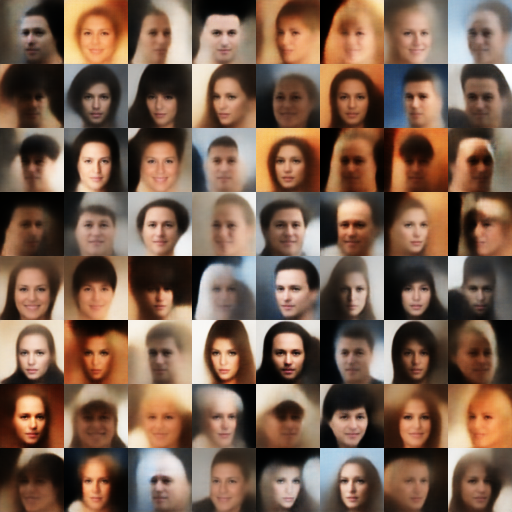}
\caption{Samples from prior; BILBO loss with $\bm{\Sigma} = \mathbf{I}/100$.}
\end{subfigure}
\caption{
Generative samples for CelebA for different methods of determining posterior variances and sample distributions. (a) Standard VAE practice using ELBO loss and sampling the standard normal prior. (b) ELBO loss but sampling from the optimal normal prior. (c) BILBO loss with constant posterior variance $\bm{\Sigma} = \mathbf{I}$. (d) BILBO loss with constant posterior variance $\bm{\Sigma} = \mathbf{I}/100$. All methods produce comparable ELBO. Sampling the standard normal prior (a) exhibits less variation than the optimal normal prior (b) which is qualitatively more similar to the variation in the reconstructions (\cref{fig-celeb-recon}). Due to the significant variation in the data, unit posterior variance (c) produces a large optimal prior variance which appears to slightly exaggerate the samples, compared to a smaller constant posterior variance (d) which is qualitatively comparable to (b).
}\label{fig-celeb-gen}
\vspace{-0.25cm}
\end{figure}

\end{document}